\documentclass[10pt,twocolumn,letterpaper]{article}

\usepackage{cvpr}              
\usepackage{booktabs}
\usepackage{multirow}
\usepackage[table,dvipsnames]{xcolor}
\usepackage{bbding}

\usepackage{marvosym}
\usepackage{lipsum}
\usepackage{pifont}

\definecolor{cvprblue}{rgb}{0.21,0.49,0.74}
\usepackage[pagebackref,breaklinks,colorlinks,allcolors=cvprblue]{hyperref}

\title{Streaming Video Instruction Tuning}

\author{%
  Jiaer Xia$^{1\ast}$ \quad Peixian Chen$^{2\ast}$ \quad  Mengdan Zhang$^2$ \quad  Xing Sun$^2$ \quad  Kaiyang Zhou$^1$\textsuperscript{\Letter} \\
  $^1$ Hong Kong Baptist University \\
  $^2$ Tencent Youtu Lab\\
  \url{https://jiaerxia.github.io/Streamo/}\\
}

\newcommand{\MethodName}{Streamo}
\newcommand{\modelname}{\texttt{Streamo}}

\begin{document}
\maketitle
\renewcommand{\thefootnote}{$\ast$}
\footnotetext[1]{Equal contribution \quad \quad
  \renewcommand{\thefootnote}{\Letter}%
  \footnotemark[1]%
  Corresponding author}
\setcounter{footnote}{0}   
\renewcommand{\thefootnote}{\arabic{footnote}}

\begin{abstract}
We present \textbf{Streamo}, a real-time streaming video LLM that serves as a general-purpose interactive assistant. Unlike existing online video models that focus narrowly on question answering or captioning, Streamo performs a broad spectrum of streaming video tasks, including real-time narration, action understanding, event captioning, temporal event grounding, and time-sensitive question answering. To develop such versatility, we construct Streamo-Instruct-465K, a large-scale instruction-following dataset tailored for streaming video understanding. The dataset covers diverse temporal contexts and multi-task supervision, enabling unified training across heterogeneous streaming tasks. After training end-to-end on the instruction-following dataset through a streamlined pipeline, Streamo exhibits strong temporal reasoning, responsive interaction, and broad generalization across a variety of streaming benchmarks. Extensive experiments show that Streamo bridges the gap between offline video perception models and real-time multimodal assistants, making a step toward unified, intelligent video understanding in continuous video streams.
\end{abstract}    
\section{Introduction}
\label{sec:intro}

Recent advances in video large language models (LLMs)~\cite{video-llama, video-llava, internvideo, bain2021frozen} have demonstrated remarkable capabilities in analyzing complete, pre-recorded videos, which establish strong baselines for offline video understanding. These models excel at holistic reasoning over long temporal sequences when given static, temporally bounded inputs~\cite{zhao2023streaming, streaming-state}, enabling tasks such as video captioning, summarization, and question answering. However, the requirements of real-time interactive AI assistants are fundamentally different: they must process continuous, unbounded video streams and respond to dynamic instructions as events unfold, often under strict latency constraints.

Existing offline models struggle to meet the demands of the streaming setting because they are designed to process entire clips before producing a single output~\cite{wang2025streambridge, qian2025dispider, wang2024mmduet}. In contrast, real-time applications require the model to continuously interpret an incoming video stream, detect when the visual context satisfies a task condition, and decide what information to output at that moment. This introduces two key challenges: 1) handling continuous, unbounded data flow without losing context, and 2) managing variable response timing and granularity across multiple tasks, which may require frame-level or longer-term temporal reasoning. A truly capable streaming video LLM must therefore integrate both task understanding and frame-level decision-making, enabling it to evaluate evolving visual contexts, determine appropriate moments to respond, and generate coherent outputs without delaying or missing critical information.

\begin{figure*}[t]
  \centering
  \includegraphics[width=1.0\linewidth]{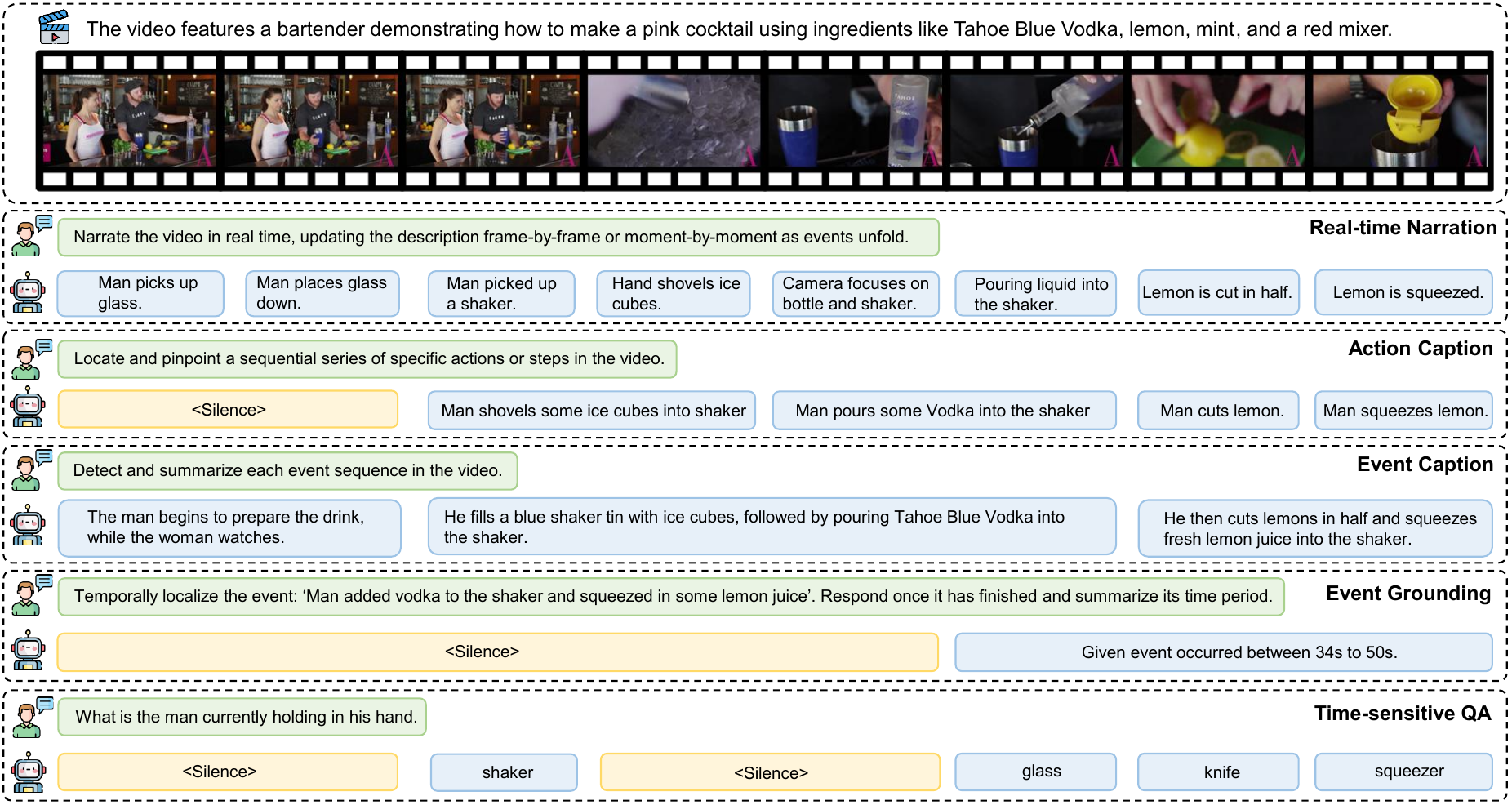}
  \caption{\textbf{An example of multi-task annotation in Streamo-Instruct-465K.} Each task is carefully labeled with the corresponding response time boundaries and content, following established annotation standards. The same video is annotated with multiple distinct tasks. The video shown in this example is sourced from ActivityNet~\cite{caba2015activitynet}.}
  \label{fig:intro}
\end{figure*}

To address these challenges, recent studies~\cite{wang2025streambridge, qian2025dispider, wang2024mmduet} have attempted to extend offline video models for streaming by introducing a separate decision module that predicts response states before invoking the offline model to generate content. While this approach preserves the reasoning capacity of the base model, it creates a trade-off between accuracy and efficiency: lightweight decision modules often lack the capacity to fully understand complex instructions and temporal dependencies, while larger modules substantially increase computational cost and inference latency. Moreover, separating decision-making from response generation prevents tight coupling between perception and response, limiting the model’s ability to seamlessly adapt to rapidly changing streaming contexts.

In this work, we propose \modelname{}\footnote{The letter \texttt{o} in \modelname{} means `omni', reflecting its multi-task and multi-modal capabilities.}, a real-time streaming video LLM that unifies decision-making and response generation in an end-to-end manner. Instead of relying on an external controller, we embed frame-level response state prediction directly into the model. Specifically, three decision heads---\textit{Silence}, \textit{Standby}, and \textit{Response}---allow the model to continuously monitor the input stream and make fine-grained judgments about when to output. Once a response state is triggered, the model immediately produces the corresponding textual output, achieving one-pass inference that significantly improves both the accuracy of response timing and the efficiency of real-time generation.

Training \modelname{} requires high-quality, temporally consistent supervision, yet existing datasets often combine heterogeneous sources with inconsistent annotation standards~\cite{han2023shot2story20k, gao2017charades, huang2020vitt}. These inconsistencies make it difficult for the model to learn precise temporal alignment or multi-task response behaviors. To overcome this problem, we construct \textbf{Streamo-Instruct-465K}, a large-scale, multi-task instruction-following dataset designed specifically for streaming video understanding and interaction. The dataset standardizes three levels of response granualarity, provides unified temporal annotations for event boundaries, and covers diverse tasks including real-time narration, action and event captioning, temporal grounding, and time-sensitive question answering. Each video is annotated for multiple tasks, providing consistent guidance that strengthens both instruction-following and temporal reasoning. An example of the annotations is shown in Fig.~\ref{fig:intro}.

Extensive experiments demonstrate that our end-to-end training paradigm effectively converts offline models into online streaming assistants. \modelname{} outperforms existing online approaches across both streaming and offline benchmarks, exhibiting strong temporal awareness, accurate frame-level decision-making, and robust multi-task instruction-following. To further support research in this domain, we also introduce a comprehensive streaming benchmark named \textbf{Streamo-Bench}, which evaluates instruction understanding across diverse interactive tasks.

Our contributions are threehold: 1) We propose a simple and effective end-to-end training framework that converts offline video models into real-time straeming assistants. 2) We introduce a multi-task instruction tuning dataset with unified temporal annotation and fine-grained response supervision. To our knowledge, this is the largest scale instruction tuning dataset for streaming video understanding and interaction. 3) We establish a comprehensive benchmark for streaming video instruction-following and provide strong baseline models for future research. All research resources including code, models, and datasets will be made publicly available.
\section{Related Work}
\label{sec:related}

\noindent
\textbf{Video Large Language Models}\quad
The field of vision foundation models~\cite{liu2023llava, videogpt+, liu2024oryx, chen2024internvl} has made remarkable progress in recent years, extending capabilities from static image understanding to more general video comprehension. Building on this foundation, numerous advanced video LLMs have emerged. For example, InternVideo2.5~\cite{internvideo25} can process videos spanning several hours, while Keye-VL-1.5~\cite{yang2025kwai} demonstrates sophisticated reasoning abilities, effectively performing complex thinking process based on video content. A critical limitation, however, is that these state-of-the-art models operate in an offline fashion, requiring the entire video as input before producing any output. This single-pass approach prevents them from handling continuous video streams, as they lack mechanisms to identify the precise temporal moments for generating responses in ongoing streams.

\noindent
\textbf{Streaming Video Understanding}\quad
To tackle real-time interaction, various methods have been proposed in the literature to turn offline video LLMs into online assistants that can identify the appropriate moment to respond in video streams. For instance, Dispider~\cite{qian2025dispider} and StreamBridge~\cite{wang2025streambridge} employ an auxiliary model to segment a video stream into fixed-length clips before feeding them to an offline model. However, this strategy introduces significant computational overhead in both training and inference and often fails to maintain context during multi-turn interactions. On the other hand, VideoLLM-Online~\cite{videollm-online} and StreamingVLM~\cite{xu2025streamingvlm} train the model in a supervised way to directly predict response timing using a special \textit{[EOS]} token. However, this approach is limited to real-time narration and cannot balance between silence and response state. To overcome these problems, we propose an end-to-end training framework along with a multi-task instruction-following dataset specifically designed for streaming video understanding and interaction.

\noindent
\textbf{Streaming Video Benchmarks}\quad
OVO-Bench~\cite{li2025ovobench} introduces 12 distinct tasks, incorporating tests for a model's ability to proactively respond. Similarly, STREAMBENCH~\cite{xiong2025streambench} and SVBENCH~\cite{yang2025svbench} concentrate on assessing multi-turn conversational abilities within continuous video contexts. A key limitation, however, is their predominant reliance on question-answer (QA) style setups—typically requiring the model to choose an answer from given options—which does not adequately assess broader instruction-following abilities such as event grounding and captioning. Motivated by the goal that streaming video models should evolve into real-time AI assistants, we introduce Streamo-Bench, a benchmark designed to probe a model’s perceptual and responsive capabilities across diverse instructions, moving beyond the constraints of traditional QA-based evaluation.
\section{Streamo: Architecture and Training}
\label{sec:method}

\subsection{Preliminaries}
Traditional video understanding models~\cite{bai2023qwen, chen2024sharegpt4video} follow an offline paradigm where the complete video $V$, question $Q$, and answer $A$ are processed using a single-turn format. Formally, given a video $V=\{v_1,v_2,...,v_T\}$ of length $T$ and a question $Q$, the model directly generates an answer $A$. This approach assumes that the entire video is accessible before inference begins, which is impractical for real-time streaming scenarios where video frames arrive sequentially.

In contrast to offline settings, streaming video understanding processes video content as it arrives in a continuous stream. The model must make decisions based on partial observations $V_{:t}={v_1,v_2,...,v_t}$, where $t \leq T_t$ meaning that the model does not have access to future frames. This temporal constraint requires fundamental changes to both the data structure and training paradigm.

\subsection{Data Structure}
To simulate streaming scenarios during training, we reformulate the single-turn offline format into a multi-turn dialogue structure. Specifically, a complete video $V$ is temporally segmented into $N$ contiguous segments:
\begin{equation}
    V = \{V^{(1)}, V^{(2)}, ..., V^{(N)}\}
    \label{eq:video}
\end{equation}
where $V^{(i)}$ denotes the $i$-th video segment. Each segment is explicitly annotated with temporal boundaries using special markers, \eg, \textit{\textless 2s-3s\textgreater}, to encode temporal information.
The multi-turn dialogue is constructed as:
\begin{equation}
    \mathcal{D} = \{(V^{(1)}, R^{(1)}), (V^{(2)}, R^{(2)}), ..., (V^{(N)}, R^{(N)})\}    
    \label{eq:dialogue}
\end{equation}
where $R^{(i)}$ denotes the response at turn $i$. Questions and answers are strategically inserted at appropriate turns based on the dataset characteristics and task requirements.

To enable efficient parallel training while maintaining compatibility with standard supervised fine-tuning paradigms, we convert decision process into predictions for the following state tokens:
\begin{itemize}
    \renewcommand{\labelitemi}{}
    \item \textit{\textless Silence\textgreater}: The model remains silent and continues processing incoming frames.
    \item \textit{\textless Standby\textgreater}: The model detects relevant video input and waits for complete information.
    \item \textit{\textless Response\textgreater}: The model receives enough information and will generate a response.
\end{itemize}

\begin{figure}[t]
  \centering
   \includegraphics[width=1.0\linewidth]{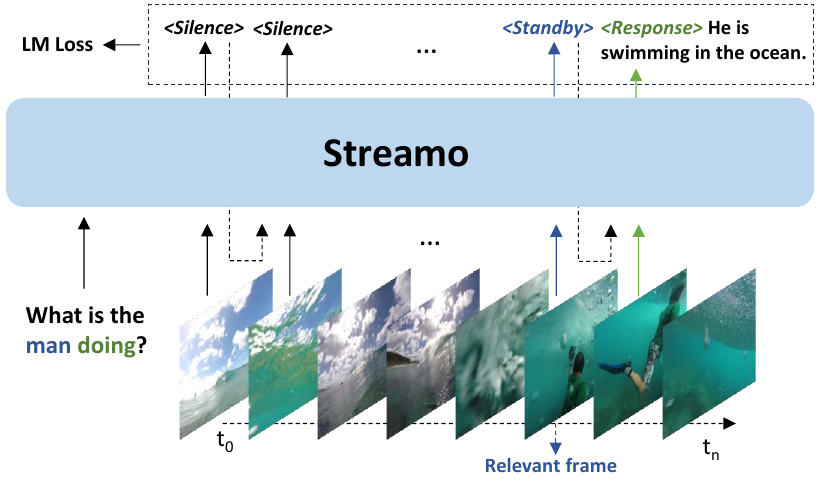}
   \caption{\textbf{Streamo's architecture.} Streaming video data is organized into an interleaved, multi-turn dialogue structure that directly integrates a response-state token into the data sequence, enabling end-to-end parallel training.}
   \label{fig:method}
\end{figure}

This design empowers the model with frame-level decision-making capabilities while maintaining the next-token prediction framework. As illustrated in Fig.~\ref{fig:method}, three discrete response states are directly integrated into the normal token prediction process: the model outputs \textit{\textless Standby\textgreater} upon detecting relevant input and \textit{\textless Response\textgreater} when it is ready to answer. A training example is shown in Tab.~\ref{tab:dialogue}. With this multi-turn dialogue training format, we can simulate realistic streaming video interactions and pose questions at any point in time.

\begin{table}[t]
\centering
\caption{The format of a multi-turn dialogue.}
\label{tab:dialogue}
\resizebox{\linewidth}{!}{
\begin{tabular}{@{}lll@{}}
\toprule
\multicolumn{3}{@{}l}{{SYSTEM PROMPT}}                                                              \\ \midrule
USER                  & \multicolumn{2}{l}{\textit{\textless 0s-1s\textgreater \textless video\textgreater}}                                            \\
ASSISTANT             & \multicolumn{2}{l}{\textit{\textless Silence\textgreater}}                                                  \\ \midrule
\multirow{2}{*}{USER} & \multicolumn{2}{l}{\textit{\textless 1s-2s\textgreater \textless video\textgreater}}                                            \\
                      & \multicolumn{2}{l}{{Notify me when the light turns green.}}               \\
ASSISTANT             & \multicolumn{2}{l}{\textit{\textless Silence\textgreater}}                                                  \\ \midrule
USER                  & \multicolumn{2}{l}{\textit{\textless 2s-3s\textgreater \textless video\textgreater}}                                            \\
ASSISTANT             & \multicolumn{2}{l}{\textit{\textless Silence\textgreater}}                                                  \\ \midrule
USER                  & \multicolumn{2}{l}{\textit{\textless 3s-4s\textgreater \textless video\textgreater}}                                            \\
ASSISTANT             & \multicolumn{2}{l}{\textit{\textless Standby\textgreater}}                                                  \\ \midrule
USER                  & \multicolumn{2}{l}{\textit{\textless 4s-5s\textgreater \textless video\textgreater}}                                            \\
ASSISTANT             & \multicolumn{2}{l}{\textit{\textless Response\textgreater} {The light just turned green.}}           \\ \bottomrule
\end{tabular}%
}
\end{table}

\subsection{Training}
The multi-turn streaming format introduces severe class imbalance among the three response states. In typical streaming scenarios, \textit{\textless Silence\textgreater} tokens dominate the distribution (often more than 80\% of the time), while \textit{\textless Response\textgreater} tokens are sparse. This imbalance biases the model toward remaining silent, making it difficult to learn response timing.

To mitigate this, we apply focal weighting~\cite{lin2017focal} specifically to the three special state tokens. Let $\mathcal{S} = \{s_{\text{silence}}, s_{\text{standby}}, s_{\text{response}}\}$ denote the special token for the three states. 
For each prediction, we compute a focal weight that emphasizes hard examples:
\begin{equation}
    w_{\text{focal}}(x_i) = (1 - p_{c_i})^{\gamma},
    \label{loss:focal}
\end{equation}
where $x_i$ represents the input features at position $i$, and $p_{c_i}$is the predicted probability for the true class $c_i$ at position $i$. $\gamma \geq 0$ is the focusing parameter that controls the rate at which easy examples are down-weighted. To further balance the rare classes, we introduce frequency-based alpha weights. For each special token $k \in \mathcal{S}$ with count $n_k$ in the current batch:
\begin{equation}
    \alpha_k = \frac{1}{|\mathcal{S}|} \cdot \frac{\sum_{j \in \mathcal{S}} n_j}{n_k},
    \label{loss:token}
\end{equation}
where $|\mathcal{S}| = 3$ is the number of special states. This assigns larger weights to less frequent special tokens.

The final loss combines the focal weighting and frequency balancing:
\begin{equation}
\mathcal{L}_i=
\begin{cases}
\alpha_{t_i}\,w_{\text{focal}}(i)\,\mathcal{L}_{\text{CE}}(i,t_i), & t_i\in\mathcal{S}\\[2mm]
\mathcal{L}_{\text{CE}}(i,t_i), & \text{otherwise}
\end{cases},
    \label{loss:final}
\end{equation}
The two weighting mechanisms are computed independently and multiplied into the cross-entropy loss. Together, they focus the model on both challenging and infrequent tokens, improving learning of response timing despite severe class imbalance in streaming data. The $\mathcal{L}_{\text{CE}}$ is the standard cross-entropy loss:
\begin{equation}
    \mathcal{L}_{\text{CE}}(i,t_i)=-\log p_{t_i}=\log\sum_{j=1}^{|\mathcal{V}|}e^{z_{i,j}}-z_{i,t_i},
    \label{loss:ce}
\end{equation}
where $z_{i,j}$ is the logit for token $j$ at position $i$ and $|\mathcal{V}|$ is the vocabulary size. This computes the negative log-likelihood of the true token. The total loss averages over all valid (non-masked) positions indicated by $\mathcal{M}$:
\begin{equation}
    \mathcal{L}_{\text{total}}=\frac{1}{|\mathcal{M}|}\sum_{i\in\mathcal{M}}\mathcal{L}_i.
    \label{loss:avg}
\end{equation}
This ensures that the loss is not affected by sequence length variations across examples in the batch.

\section{Streamo-Instruct-465K}
\label{sec:streaminstruct}

\begin{figure*}[t]
  \centering
  \begin{subfigure}{0.4\linewidth}
    \includegraphics[width=\linewidth]{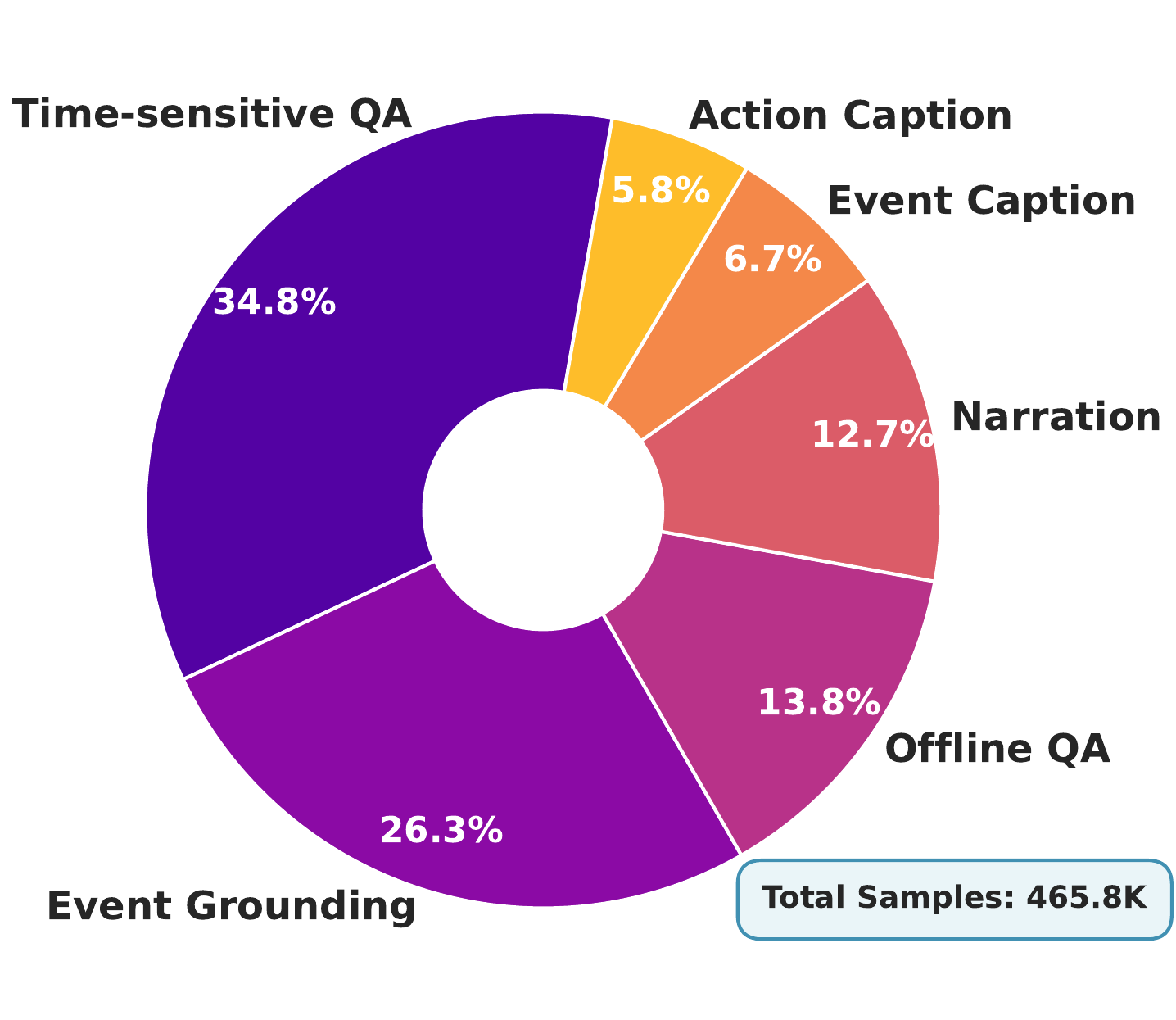}
    \label{fig:short-a}
  \end{subfigure}
  \hfill
  \begin{subfigure}{0.55\linewidth}
    \includegraphics[width=\linewidth]{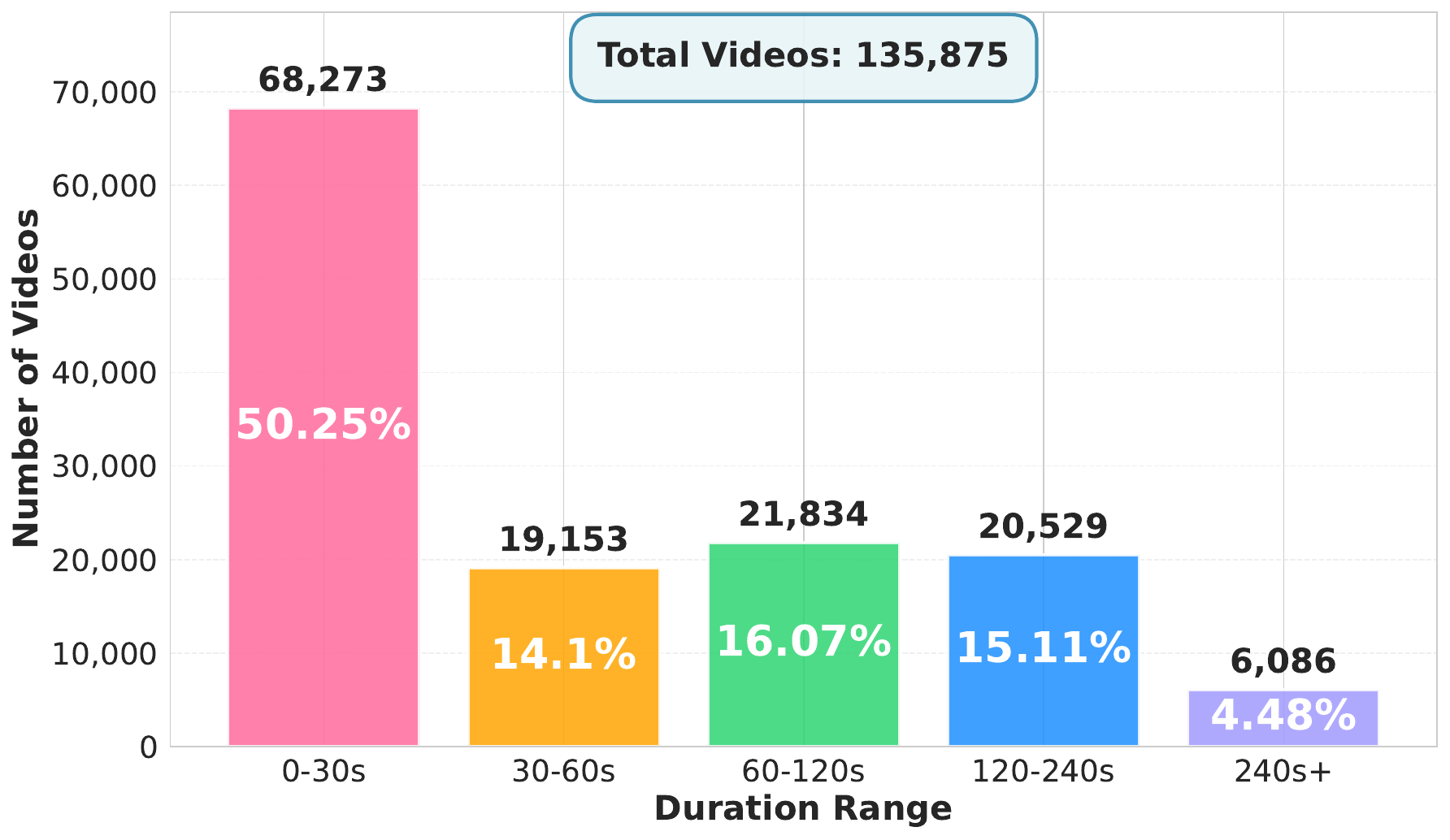}
    \label{fig:short-b}
  \end{subfigure}
  \caption{Dataset distribution overview. \textbf{Left}: task distribution; \textbf{Right}: video duration distribution.}
    \label{fig:statistics}
\end{figure*}

\subsection{Data Construction}
To provide clear supervision for each round of response decisions, we re-annotated a large-scale training set with detailed temporal boundary labels based on the existing open-source video datasets. We predefined multiple tasks spanning different response granularities, assigning each video several types of task annotations. This approach offers several advantages. First, a unified annotation protocol is applied across datasets, avoiding the inconsistencies and biases that arise when naively mixing datasets with heterogeneous labeling standards. Additionally, each video carries multiple task types with clearly delineated response boundaries, enabling the model to better perceive and understand varying task requirements, develop robust instruction-following capabilities, and execute a range of real-time response tasks. Below, we detail the annotation protocol for each task.

\noindent
\textbf{Real-time Narration}\quad
This task performs real-time commentary over video, requiring second-by-second descriptions that capture fine-grained visual changes. The annotation protocol is: 1) segment each video at one-second intervals; 2) for every adjacent pair of one-second segments (i.e., a two-second window), use Qwen2.5-VL-72B~\cite{bai2025qwen25vl} to describe the changes observed between them; 3) concatenate the per-second outputs and send the full narration to GLM-4.5~\cite{zeng2025glm45} for post-processing to remove repetitions and redundancies, smooth transitions, and ensure coherent, context-aware narration.

\noindent
\textbf{Event Caption}\quad
This task is similar to standard video captioning but requires the model to detect event boundaries and provide the corresponding caption when an event ends. To construct supervision: 1) generate segment-level captions with the ARC-Hunyuan-Video-7B~\cite{ge2025arc} model; 2) temporally ground each caption using the same model; 3) retain only those videos in which all segment captions have mutually consistent, overlapping time spans that align with the original output. This yields two benefits: it filters out erroneous, noisy data and produces samples with sharper, more explicit event boundaries, enabling clearer supervision.

\noindent
\textbf{Action Caption}\quad
This task mirrors event captioning but narrows the focus from dense events to discrete actions or procedural steps. We reuse the event-caption pipeline and augment it with action-oriented prompts and targeted filtering. This produces cleaner, step-level supervision with sharper action delineation.

\noindent
\textbf{Event Grounding}\quad
The grounding annotation is similar to the offline setup, where each sample pairs an event caption with its corresponding temporal span. The key difference in the online setting is that the caption is provided in advance, and the model must continuously monitor the subsequent video stream to detect the specified event and localize its occurrence in time. We randomly sample captions from the event-caption annotations, rewrite them for grounding, and integrate existing datasets to broaden coverage and improve robustness.

\noindent
\textbf{Time-sensitive QA}\quad
This task targets questions whose correct answers change over time in a dynamic video stream. To construct supervision: 1) process each video with GLM-4.5V~\cite{v2025glm45v} model to detect change points across multiple aspects—object attributes (e.g., color, size, state), spatial positions, actions and interactions, counts, and scene or context shifts; 2) generate question–answer pairs from these variations by posing a single, unified question and providing diverse, time-specific answers at the corresponding time points.

\subsection{Statistics}
Using a unified annotation standard and protocol, we labeled and curated a total of 400K valid samples and additionally merged offline video QA data from the LLaVA-Video~\cite{llava-video} dataset, culminating in Streamo-Instruct-465K, and the task distribution is shown on the left of \cref{fig:statistics}. We integrated multiple open-source video datasets as sources, including Koala~\cite{wang2025koala}, LLaVA-Video~\cite{llava-video}, ActivityNet~\cite{caba2015activitynet}, QVHighlight~\cite{qvhighlight}, YouCook2~\cite{zhou2018youcook2}, HACS~\cite{zhao2019hacs}, EgoTimeQA~\cite{di2024egotimeqa}, DiDeMo~\cite{didemo}, and COIN~\cite{tang2019coin}, yielding 135,875 videos in total. The distribution of video durations is shown on the right of \cref{fig:statistics}.

\section{Experiments}
\label{sec:exp}

\begin{table*}[t]
\centering
\setlength{\tabcolsep}{3pt}
\renewcommand{\arraystretch}{1.2}
\caption{Comparison with state-of-the-art on OVO-Bench. `Streamo Framework' denotes adapting offline models to the online setting using our training framework. ET-Instruct-3B is trained with ET-Instruct-164K and $^\dagger$ indicates LLaVA-Video data is added as offline support. $^*$ means the model is trained at 1 fps and evaluated at 2 fps.}
\label{tab:main_exp}
\resizebox{\textwidth}{!}{%
\begin{tabular}{@{}lccccccccccccccccc@{}}
\toprule
\multicolumn{1}{l|}{\multirow{2}{*}{Model}} & \multicolumn{1}{c|}{\multirow{2}{*}{\# Frames}} & \multicolumn{7}{c|}{Real-Time Visual Perception}                                                & \multicolumn{4}{c|}{Backward Tracing}                                   & \multicolumn{4}{c|}{Forward Active Responding}                          & Overall Avg. \\ \cmidrule(l){3-18} 
\multicolumn{1}{l|}{}                       & \multicolumn{1}{c|}{}                           & OCR   & ACR   & ATR   & STU   & FPD   & \multicolumn{1}{c|}{OJR}   & \multicolumn{1}{c|}{Avg.}  & EPM   & ASI   & \multicolumn{1}{c|}{HLD}   & \multicolumn{1}{c|}{Avg.}  & REC   & SSR   & \multicolumn{1}{c|}{CRR}   & \multicolumn{1}{c|}{Avg.}  & Overall Avg. \\ \midrule
\multicolumn{18}{c}{\textbf{Open-source Offline Models}} \\ \midrule
\multicolumn{1}{l|}{Qwen2-VL-72B~\cite{qwen2vl}}           & \multicolumn{1}{c|}{64}                         & 65.77 & \textbf{60.55} & \textbf{}69.83 & 51.69 & 69.31 & \multicolumn{1}{c|}{54.35} & \multicolumn{1}{c|}{61.92} & 52.53 & \textbf{60.81} & \multicolumn{1}{c|}{\textbf{57.53}} & \multicolumn{1}{c|}{\textbf{56.95}} & \textbf{38.83} & 64.07 & \multicolumn{1}{c|}{45}    & \multicolumn{1}{c|}{49.3}  & \textbf{56.27}        \\
\multicolumn{1}{l|}{LLaVA-Video-7B~\cite{llava-video}}         & \multicolumn{1}{c|}{64}                         & \textbf{69.13} & 58.72 & 68.83 & 49.44 & \textbf{74.26} & \multicolumn{1}{c|}{59.78} & \multicolumn{1}{c|}{63.52} & \textbf{56.23} & 57.43 & \multicolumn{1}{c|}{7.53}  & \multicolumn{1}{c|}{40.4}  & 34.1  & \textbf{69.95} & \multicolumn{1}{c|}{60.42} & \multicolumn{1}{c|}{\textbf{54.82}} & 52.91        \\
\multicolumn{1}{l|}{LLaVA-OneVision-7B~\cite{llava-onevision}}     & \multicolumn{1}{c|}{64}                         & 66.44 & 57.8  & \textbf{73.28} & \textbf{53.37} & 71.29 & \multicolumn{1}{c|}{\textbf{61.96}} & \multicolumn{1}{c|}{\textbf{64.02}} & 54.21 & 55.41 & \multicolumn{1}{c|}{21.51} & \multicolumn{1}{c|}{43.71} & 25.64 & 67.09 & \multicolumn{1}{c|}{58.75} & \multicolumn{1}{c|}{50.5}  & 52.74        \\
\multicolumn{1}{l|}{Qwen2-VL-7B~\cite{qwen2vl}}            & \multicolumn{1}{c|}{64}                         & 60.4  & 50.46 & 56.03 & 47.19 & 66.34 & \multicolumn{1}{c|}{55.43} & \multicolumn{1}{c|}{55.98} & 47.81 & 35.48 & \multicolumn{1}{c|}{56.08} & \multicolumn{1}{c|}{46.46} & 31.66 & 65.82 & \multicolumn{1}{c|}{48.75} & \multicolumn{1}{c|}{48.74} & 50.39        \\
\multicolumn{1}{l|}{InternVL-V2-8B~\cite{internvl}}         & \multicolumn{1}{c|}{64}                         & 67.11 & \textbf{60.55} & 63.79 & 46.07 & 68.32 & \multicolumn{1}{c|}{56.52} & \multicolumn{1}{c|}{60.39} & 48.15 & 57.43 & \multicolumn{1}{c|}{24.73} & \multicolumn{1}{c|}{43.44} & 26.5  & 59.14 & \multicolumn{1}{c|}{54.14} & \multicolumn{1}{c|}{46.6}  & 50.15        \\
\multicolumn{1}{l|}{LongVU-7B~\cite{longvu}}              & \multicolumn{1}{c|}{1fps}                       & 53.69 & 53.21 & 62.93 & 47.75 & 68.32 & \multicolumn{1}{c|}{59.78} & \multicolumn{1}{c|}{57.61} & 40.74 & 59.46 & \multicolumn{1}{c|}{4.84}  & \multicolumn{1}{c|}{35.01} & 12.18 & 69.48 & \multicolumn{1}{c|}{\textbf{60.83}} & \multicolumn{1}{c|}{47.5}  & 46.71        \\ \midrule
\multicolumn{18}{c}{\textbf{Open-source Online Models}}  \\ \midrule
\multicolumn{1}{l|}{Flash-VStream-7B~\cite{zhang2024flashvstream}}       & \multicolumn{1}{c|}{1fps}                       & 24.16 & 29.36 & 28.45 & 33.71 & 25.74 & \multicolumn{1}{c|}{28.8}  & \multicolumn{1}{c|}{28.37} & 39.06 & 37.16 & \multicolumn{1}{c|}{5.91}  & \multicolumn{1}{c|}{27.38} & 8.02  & 67.25 & \multicolumn{1}{c|}{60}    & \multicolumn{1}{c|}{45.09} & 33.61        \\
\multicolumn{1}{l|}{VideoLLM-online-8B~\cite{videollm-online}}     & \multicolumn{1}{c|}{2fps}                       & 8.05  & 23.85 & 12.07 & 14.04 & 45.54 & \multicolumn{1}{c|}{21.2}  & \multicolumn{1}{c|}{20.79} & 22.22 & 18.8  & \multicolumn{1}{c|}{12.18} & \multicolumn{1}{c|}{17.73} & -     & -     & \multicolumn{1}{c|}{-}     & \multicolumn{1}{c|}{-}     & -            \\
\multicolumn{1}{l|}{Dispider-7B~\cite{qian2025dispider}}               & \multicolumn{1}{c|}{1fps}                       & 57.72 & 49.54 & 62.07 & 44.94 & 61.39 & \multicolumn{1}{c|}{51.63} & \multicolumn{1}{c|}{54.55} & 48.48 & 55.41 & \multicolumn{1}{c|}{4.3}   & \multicolumn{1}{c|}{36.06} & 18.05 & 37.36 & \multicolumn{1}{c|}{48.75} & \multicolumn{1}{c|}{34.72} & 41.78        \\ 

\multicolumn{1}{l|}{ViSpeak-7B~\cite{vispeak}}               & \multicolumn{1}{c|}{1fps}                       & \textbf{75.17} & \textbf{58.72} & \textbf{71.55} & \textbf{51.12} & \textbf{74.26} & \multicolumn{1}{c|}{\textbf{66.85}} & \multicolumn{1}{c|}{\textbf{66.28}} & \textbf{59.93} & \textbf{48.65} & \multicolumn{1}{c|}{\textbf{63.98}}   & \multicolumn{1}{c|}{\textbf{57.52}} & \textbf{33.81} & \textbf{68.52} & \multicolumn{1}{c|}{\textbf{60.42}} & \multicolumn{1}{c|}{\textbf{54.25}} & \textbf{61.08}       \\ 
\midrule
\multicolumn{18}{c}{\textbf{Streamo Framework}}                                                                                                                                          \\ \midrule
\multicolumn{1}{l|}{ET-Instruct-3B~\cite{et-instruct}}    & \multicolumn{1}{c|}{1fps}                       & 65.10 & 35.78  & 56.90  &  35.39  &  24.75  & \multicolumn{1}{c|}{60.87}      & \multicolumn{1}{c|}{46.47}      &   41.81    &   35.14    & \multicolumn{1}{c|}{8.6}      & \multicolumn{1}{c|}{28.52}      &   20.06    &  52.31     & \multicolumn{1}{c|}{67.50}      & \multicolumn{1}{c|}{46.62}      &  40.54            \\
\multicolumn{1}{l|}{ET-Instruct-3B$^{\dagger}$~\cite{et-instruct}} & \multicolumn{1}{c|}{1fps}        &  71.14     &  50.46     &  67.24     &   37.08    &  60.40     & \multicolumn{1}{c|}{60.33}      & \multicolumn{1}{c|}{57.78}      &  48.82     & 48.56     & \multicolumn{1}{c|}{11.29}      & \multicolumn{1}{c|}{36.22}      & 13.68      &   48.62    & \multicolumn{1}{c|}{60.00}      & \multicolumn{1}{c|}{40.77}      &   44.92           \\
\rowcolor{SkyBlue!20}\multicolumn{1}{l|}{\MethodName-3B}    & \multicolumn{1}{c|}{1fps}                       & 78.52 & 52.29 & 67.24 & 44.38 & 55.45 & \multicolumn{1}{c|}{71.20}  & \multicolumn{1}{c|}{61.51} & 51.18 & 57.43 & \multicolumn{1}{c|}{16.67} & \multicolumn{1}{c|}{41.76} & 27.94 & 50.72 & \multicolumn{1}{c|}{82.5}  & \multicolumn{1}{c|}{53.72} & 52.33        \\ \midrule
\rowcolor{SkyBlue!20}\multicolumn{1}{l|}{\MethodName-7B}    & \multicolumn{1}{c|}{1fps}                       & \textbf{79.19} & 57.80  & 75.00    & \textbf{49.44} & 64.36 & \multicolumn{1}{c|}{70.11} & \multicolumn{1}{c|}{65.98} & 54.55 & 52.03 & \multicolumn{1}{c|}{31.72} & \multicolumn{1}{c|}{46.10} & 29.96 & 51.03 & \multicolumn{1}{c|}{\textbf{83.33}} & \multicolumn{1}{c|}{54.77} & 55.61        \\
\rowcolor{SkyBlue!20}
\multicolumn{1}{l|}{\MethodName-7B}    & \multicolumn{1}{c|}{\ 2fps$^*$}                       & 77.18 & \textbf{66.06} & \textbf{76.72} & 45.51 & \textbf{66.34} & \multicolumn{1}{c|}{\textbf{72.83}} & \multicolumn{1}{c|}{\textbf{67.44}} & \textbf{55.56} & \textbf{58.11} & \multicolumn{1}{c|}{\textbf{33.87}} & \multicolumn{1}{c|}{\textbf{49.18}} & \textbf{30.84} & \textbf{57.55} & \multicolumn{1}{c|}{82.5}  & \multicolumn{1}{c|}{\textbf{56.96}} & \textbf{57.86 }       \\ \bottomrule
\end{tabular}%
}
\end{table*}

\subsection{Models and Datasets}
To assess the effectiveness of our training strategy, we adopt Qwen2.5-VL~\cite{bai2025qwen25vl} as our base model, across both 3B and 7B model size. Meanwhile, we additionally conduct experiments based on several existing state-of-the-art offline models, including Qwen3-VL~\cite{qwen3technicalreport}, and InternVL-3~\cite{zhu2025internvl3}, to demonstrate the compatibility of our framework; these results are presented in the Supplementary material. In addition to training on our proposed Streamo-Instruct-465K dataset, we also compare against ET-Instruct-164K~\cite{et-instruct}, a large-scale instruction-tuning dataset with rich temporal information that has been widely used in prior work to train online video models. To enable a fairer comparison with Streamo-Instruct-465K, we also report results on a mixed dataset comprising ET-Instruct-164K and LLaVA-Video.

\subsection{Benchmarks}
We evaluated our model across three dimensions of benchmarks: Online, Offline, and Stream Instruction. For the online setting, we adopted OVO-Bench~\cite{li2025ovobench}, which covers three temporal perception modes, including real-time, backward, and forward, and also spans a total of 12 subtasks. The offline evaluation used standard general video understanding benchmarks, including the short-video benchmarks MVBench~\cite{mvbench} and TempCompass~\cite{liu2024tempcompass}, as well as the long-video benchmarks VideoMME~\cite{videomme} and LongVideoBench~\cite{wu2024longvideobench}, providing a comprehensive assessment of capabilities. In addition, to assess multi-instruction following in an online context, we constructed Streamo-Bench, which includes 300 videos and 3,000 instruction tasks. Each video is paired with tasks of varying temporal scopes and granularities to measure the model’s adherence to instructions, providing an important metric for building a reliable real-time AI assistant. Detailed information for Streamo-Bench is given in the Supplementary material.

\subsection{Implementation Details}
Across all models, we use a unified training setup. Full parameter tuning is applied with the vision encoder frozen, and only the connector and the LLM will be updated. Training runs for a single epoch with a batch size of 512 and a learning rate of 1$e$-5. For multi-turn dialogue construction, each video is split into turns of one second, and frames are sampled at 1 fps. The hyperparameter gamma in \cref{loss:focal} is set to 2. In experiments that include LLaVA-Video, we restrict the training data to the same subset used by Streamo-Instruct-465K to ensure a direct and fair comparison.

\begin{table*}[t]
\centering
\setlength{\tabcolsep}{4pt}
\renewcommand{\arraystretch}{1.1}
\caption{Results on offline video benchmarks. The table compares converted online models with their original offline base models and SOTA models. Numbers in parentheses denote performance differences from the corresponding offline models.}
\label{tab:offline}
\resizebox{0.9\textwidth}{!}{%
\footnotesize
\begin{tabular}{@{}l|cccccc|c@{}}
\toprule
Model                           & \begin{tabular}[c]{@{}c@{}}OVO\\ Real-Time\end{tabular} & \begin{tabular}[c]{@{}c@{}}OVO\\ Backward\end{tabular} & MVBench & TempCompass & VideoMME & LongVideoBench & Avg \\ \midrule
\multicolumn{8}{c}{\textbf{Proprietary Models}}  \\ \midrule
Gemini-1.5-pro~\cite{team2024gemini} & 69.3 & 62.5 &	60.5	&	67.1	&	75.0	&	64.0 & 66.4 \\
GPT-4o~\cite{hurst2024gpt4o}	&     64.5 & 60.8   & 64.6	&	70.9	&	71.9	& 66.7	& 66.6 \\ \midrule
\multicolumn{8}{c}{\textbf{Open-source Online Models}}  \\ \midrule
Flash-VStream-7B~\cite{zhang2024flashvstream} & 28.4      & 27.4       & 61.2    & -        & 61.2     & -     & -      \\
VideoLLM-online-8B~\cite{videollm-online}    & 20.8      & 17.7       & 33.9    & -        & 26.9     & -     & -      \\
Dispider-7B~\cite{qian2025dispider}    & 54.6      & 36.1       & -    & -        & 57.2     & -     & -      \\
StreamingVLM-7B~\cite{xu2025streamingvlm}  & 62.0      & -       & 69.2    & -        & 65.1     & 59.0     & -      \\ \midrule
\multicolumn{8}{c}{\textbf{Streamo Framework}}  \\ \midrule
Qwen2.5-VL-3B ~\cite{bai2025qwen25vl}    & 54.6      & 37.8       & 67.0    & 64.4        & 61.5     & 54.2     & 56.6      \\
ET-Instruct-3B~\cite{et-instruct}    & 46.5 \color[HTML]{CB0000}(-8.1)  & 28.6 \color[HTML]{CB0000}(-9.2)      & 65.8 \color[HTML]{CB0000}(-1.2)   & 60.3 \color[HTML]{CB0000}(-4.1)       & 56.6 \color[HTML]{CB0000}(-4.9)    & 51.2 \color[HTML]{CB0000}(-3.0) &    51.5 \color[HTML]{CB0000}(-5.1)    \\
ET-Instruct-3B$^\dagger$~\cite{et-instruct}    & 57.8 \color[HTML]{009901}(+3.2)  & 36.2 \color[HTML]{CB0000}(-1.6)   & 68.1 \color[HTML]{009901}(+1.1)   & 63.7 \color[HTML]{CB0000}(-0.7)       & 59.6 \color[HTML]{CB0000}(-1.9)    & 54.9 \color[HTML]{009901}(+0.7)  &  56.7 \color[HTML]{009901}(+0.1)     \\
\rowcolor{SkyBlue!20}
\MethodName-3B       & 61.5 \color[HTML]{009901}(+6.9)  & 41.8 \color[HTML]{009901}(+4.0)  & 67.9 \color[HTML]{009901}(+0.9)   & 66.2 \color[HTML]{009901}(+1.8)       & 61.8 \color[HTML]{009901}(+0.3)    & 56.2 \color[HTML]{009901}(+2.0)      &  59.2 \color[HTML]{009901}(+2.6) \\ \midrule
Qwen2.5-VL-7B ~\cite{bai2025qwen25vl}       & 58.8                & 42.2       & 69.6    & 71.7        & 65.1     & 56.0      &   60.6  \\
\rowcolor{SkyBlue!20}
\MethodName-7B   & 66.0 \color[HTML]{009901}(+7.2)               & 46.1 \color[HTML]{009901}(+3.9)      & 72.3 \color[HTML]{009901}(+2.7)   & 71.8     \color[HTML]{009901}(+0.1)   & 67.9 \color[HTML]{009901}(+2.8)    & 59.2 \color[HTML]{009901}(+3.2)    &  63.9 \color[HTML]{009901}(+3.3)   \\ \bottomrule
\end{tabular}%
}
\end{table*}

\subsection{Main Results}
\noindent
\textbf{Comparison with SOTA on Online Video Benchmarks}
The main results are shown in \cref{tab:main_exp}. Using the Streamo framework, we train the models with ET-Instruct and Streamo-Instruct datasets and compare their performance to currently available open-source offline and online models. The key findings are as follows: \textbf{1) \MethodName\ significantly outperforms SOTA.} It is clear that our proposed \MethodName-7B exceeds the previous SOTA, Dispider, by +13.83\% on average performance. Moreover, we observe that the model trained at 1 fps can be directly evaluated at 2 fps without retraining, achieving an additional +4.66\% performance improvement, indicating robust generalization to higher test-time frame rates; \textbf{2) Streamo-Instruct-465K dataset surpasses existing dataset.} Compared with the ET-Instruct-164K, our proposed Streamo-Instruct-465K delivers a comprehensive performance advantage, with +7.1\% on forward task and +11.79\% overall; \textbf{3) Offline supervision can hinder online learning.} Augmenting ET-Instruct with the offline LLaVA-Video dataset boosts real-time perceptual accuracy but compromises streaming ability, revealing a trade-off inherent to offline-only supervision. This also demonstrates that Streamo-Instruct-465K transfers effectively to online, streaming scenarios while maintaining strong offline perceptual capability.

\noindent
\textbf{Comparison with SOTA on Offline Video Benchmarks}
To evaluate the general video understanding capability of models after conversion to the online setting, we compare \MethodName\ against the SOTA method and original offline base model on a suite of general offline video benchmarks, with results reported in \cref{tab:offline}. The findings show that, after conversion, \MethodName\ retains strong perceptual performance on offline benchmarks across both short-form and long-form videos, surpassing the SOTA, StreamingVLM, in every benchmark. Meanwhile, models trained with our Streamo-Instruct-465K exhibit consistent improvements over base models, with \MethodName-7B achieves an average improvement of +3.4\% based on Qwen2.5-VL-7B. Holding architecture and training setup constant, Streamo-Instruct-465K also provides a clear advantage over alternative data recipes, outperforming ET-Instruction and LLaVA-Video by +7.8\% and +2.5\% on average, respectively. These results underscore that our training framework and data not only enable effective transformation of models for streaming video understanding but also preserve and enhance core perceptual capabilities on offline video tasks.

\begin{table}[t]
\centering
\setlength{\tabcolsep}{4pt}
\renewcommand{\arraystretch}{1.1}
\caption{Ablation study of loss functions for online training on OVO-Bench Forward Active tasks.}
\label{tab:ablation}
\resizebox{\columnwidth}{!}{%
\scriptsize
\begin{tabular}{@{}llcccc@{}}
\toprule
Base Model         & Loss Type    & REC     & SSR   & CRR    \\ \midrule
Qwen2.5-VL-3B & CrossEntropy & 6.45    & 20.99 & 41.67  \\
Qwen2.5-VL-3B & Loss Scale   & 18.62   & 41.02 & 49.17  \\
Qwen2.5-VL-3B & Focal Loss   & 27.94   & 50.72 & 82.5   \\ \midrule
InternVL3-2B  & CrossEntropy & 9.46    & 20.50 & 40.42  \\
InternVL3-2B  & Loss Scale   & 21.20   & 31.47 & 48.75   \\
InternVL3-2B  & Focal Loss   & 29.23   & 47.38 & 80.42   \\ \bottomrule
\end{tabular}%
}
\end{table}

\noindent
\textbf{Streamo-Bench}\quad
To evaluate the model’s ability to follow different instructions and perform varied tasks, we assign multiple instruction-driven tasks to a single video, including forward grounding, backward grounding, narration captions, dense captions, and time-sensitive question answering. Details, examples, and statistics for these tasks are presented in the Supplementary material.

\begin{table*}[t]
\centering
\setlength{\tabcolsep}{3pt}
\renewcommand{\arraystretch}{1.1}
\caption{Evaluation results on Streamo-Bench. Forward and backward grounding are determined by whether the query refers to a time point before or after the event period, and results are using the mIoU metric. Caption evaluation is conducted by calculating the win rate with Qwen2.5-VL-72B model. TSQA denotes Time-Sensitive QA, i.e., questions whose answers change over time.}
\label{tab:streambench}
\resizebox{0.8\textwidth}{!}{%
\footnotesize
\begin{tabular}{@{}l|cc|cc|cc|c@{}}
\toprule
\multirow{2}{*}{Model} & \multicolumn{2}{c|}{Grounding}               & \multicolumn{2}{c|}{Caption}                 & \multicolumn{2}{c|}{TSQA}                    & \multirow{2}{*}{Average} \\ \cmidrule(lr){2-7}
                                                & Forward     & Backward    & Narration   & Dence Caption   & Accuracy    & Recall      &                      \\ \midrule
Flash-VStream-7B~\cite{zhang2024flashvstream}   & 0           &  0          &  23.5       &  25.9           &  30.8       &  13.1       &   15.6                   \\
VideoLLM-online-8B~\cite{videollm-online}       & 0           &  0          &  42.0       &  6.6           &  19.6       &  7.6        &    12.6                  \\
Dispider-7B~\cite{qian2025dispider}             & 0           &  8.33       &  31.6       &  29.2           &  14.0       &  4.4        &   14.6                   \\
StreamingVLM-7B~\cite{xu2025streamingvlm}       & 0           & 0           &  68.5       &  24.0           & 11.8        & 43.1        &   24.6                   \\
\rowcolor{SkyBlue!20}
\MethodName-3B                                  & 14.7        & 27.5        &  71.4       &  68.5          & 20.1        & 65.7        &    44.7                  \\
\rowcolor{SkyBlue!20}
\MethodName-7B                                  & 29.4        & 38.3        &  75.9       &  72.8        & 51.6        & 63.9        &      55.3               \\ \bottomrule
\end{tabular}%
}
\vspace{-2mm}
\end{table*}

As shown in \cref{tab:streambench}, existing online models show deficiencies in comprehensive multi-task coverage. Our analysis indicates that these shortcomings stem largely from an inadequate ability to comprehend and follow complex instructions. For instance, removing predefined options leads to widespread failure—as the grounding results show—highlighting a vulnerability to open-ended prompts. Furthermore, in standard QA scenarios, models frequently overlook instructions to update answers as conditions change, which severely degrades recall. We probe instruction comprehension and prompt sensitivity further with additional experiments in the Supplementary material. Collectively, these observations expose a critical gap in current capabilities. In contrast, \modelname{} demonstrates robust performance across tasks, clearly exhibiting strong instruction-following ability. This outcome validates both the diagnostic power of our benchmark and the effectiveness of our method in learning generalized instruction-following capabilities.

\subsection{Ablation}
To evaluate the effectiveness of our focal loss for training the three decision states, \textit{\textless Silence\textgreater}, \textit{\textless Standby\textgreater}, and \textit{\textless Response\textgreater}, we compare it to standard cross-entropy loss. As shown in \cref{tab:ablation}, training without state-aware reweighting severely limits performance due to significant class imbalance. In the Streamo-Instruct-465K dataset, the empirical ratio of state labels is approximately \textit{\textless Silence\textgreater}:\textit{\textless Standby\textgreater}:\textit{\textless Response\textgreater} $=$ 12:3:2, which biases conventional training toward predicting Silence and suppresses actual Response predictions.

A straightforward remedy is to assign fixed class weights inversely proportional to label frequency. Specifically, we set the weights to 0.3, 1.3, and 2.0 for silence, standby, and response, respectively, to emphasize response timing. As illustrated in the line “Loss Scale” in \cref{tab:ablation}), this adjustment effectively mitigates the degradation caused by imbalance. However, fixed weighting fails to capture token-level hardness and sequence-level heterogeneity in decision-state distributions—for instance, narration tasks may contain multiple responses, whereas a QA task might include only one.

Our proposed focal loss addresses this limitation by dynamically reweighting losses based on token-level hardness and per-batch state frequency, thereby providing more adaptive supervision for response-timing decisions. Across both InternVL-3-2B and Qwen2.5-VL-3B backbones, training with the proposed focal loss consistently yields substantial improvements over both the vanilla cross-entropy and fixed-weight baselines.
\section{Conclusion}
\label{sec:conclusion}
Our work targets the advancement of streaming video by jointly addressing model training and data construction. We introduce an end-to-end training framework together with a large-scale instruction-tuning dataset, Streamo-Instruct-465K, enabling the conversion of multiple state-of-the-art offline models into online version. The resulting model, \modelname{}, not only excels on streaming benchmarks but also rivals top-performing offline models. Furthermore, our proposed Streamo-Bench, which simulates complex multi-instruction scenarios, showcases \modelname{}'s robust multi-tasking capabilities. Collectively, these contributions mark a significant leap towards creating general-purpose, real-time, and interactive AI assistants.

\section{Limitations and Future Work}
In terms of limitations, while our approach achieves strong accuracy, it is limited by the inherent challenges of streaming video's unbounded temporal context. Our current pipeline lacks specialized long-sequence optimizations, leading to significant memory and latency costs that become prohibitive as sequence length grows.

By leveraging our framework’s compatibility with existing techniques, we can integrate KV-cache management and visual token pruning to reduce computational overhead, alongside exploring sliding-window attention and adaptive frame compression for refined context management. Collectively, these strategies are designed to enhance training and inference efficiency, extend the effective context length, and facilitate an unbounded, real-time data stream.

\section{Acknowledgement}
This research is supported by Hong Kong Research Grants Council Early Career Scheme (No. 22200824).

{
    \small
    \bibliographystyle{ieeenat_fullname}
    \bibliography{main}
}
\clearpage
\appendix
\setcounter{page}{1}
\maketitlesupplementary
\section{Streamo}
\subsection{System Prompt}
We design a dedicated system prompt for \modelname{} that enables the model to handle dynamic streaming video content, interpret three predefined response states, and make real-time decisions at the frame level. The full prompt is provided in \cref{prompt:streamo}. This deliberately crafted prompt helps the model quickly adapt to the streaming input pattern and perform the required behavior transformation.

\subsection{Instruction Prompt}
In \cref{prompt:template}, we present the prompt templates used for all tasks. These diverse task instructions help the model better understand different task requirements, thereby fostering more general multi-task instruction-following capabilities. This goes beyond prior setups where models were confined to standalone QA, and represents a step toward general real-time interactive AI.

\subsection{More Experimental Results}
Our training framework converts offline models into streaming-capable models with minimal intrusive modifications, enabling these base models to process streaming video data. This design yields strong compatibility and allows direct application to a wide range of offline models. In Tab.\ref{tab:sup_online} and\ref{tab:sup_offline}, we further report results using InternVL3\cite{zhu2025internvl3} and Qwen3VL\cite{qwen3technicalreport} as \modelname’s base models. These results show that our framework effectively leverages the capabilities of offline models and extends them to online streaming video processing. This is particularly advantageous given the rapid iteration of offline models, as our framework can readily harness their improvements for real-time interactive video understanding. 

Meanwhile, we also evaluated Streamo on ViSpeak-Bench~\cite{vispeak}, shown in \cref{tab:ViSpeakBench}. The results show that our method achieves a clear advantage in response-time accuracy, demonstrating the effectiveness and soundness of our response architecture.

\subsection{Visualization}
\begin{figure*}[t]
  \centering
  \vspace*{-9mm} 
  \includegraphics[width=0.9\linewidth]{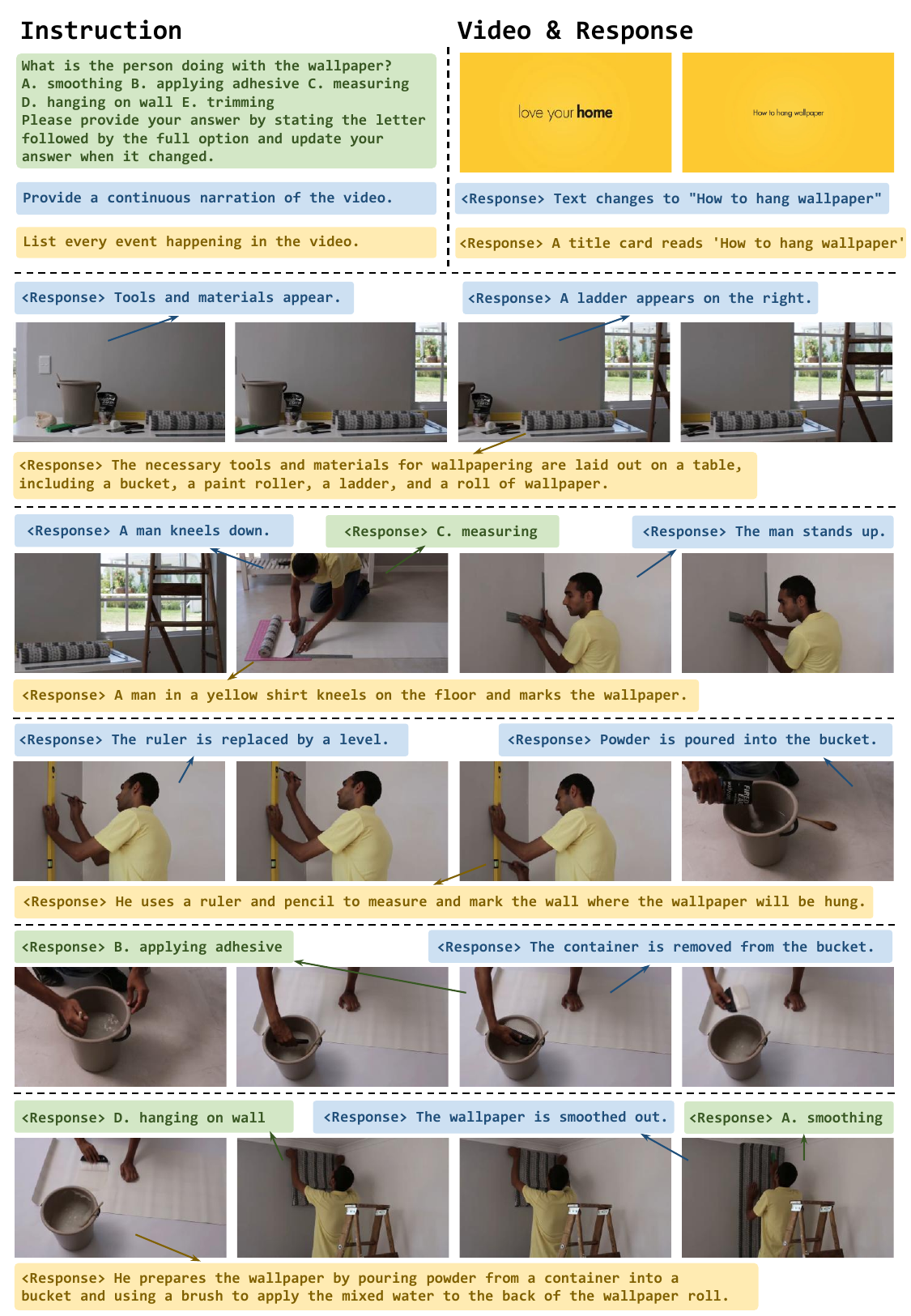}
  \caption{\textbf{Visualization of Streamo's outputs.} We present the results of Streamo processing the same video under different task instructions and distinguish them using different colors, including \textcolor{Green}{TSQA}, \textcolor{RoyalBlue}{Narration}, and \textcolor{Brown!50}{Caption}. Arrows indicate the frames corresponding to the response moment.}
  \label{fig:visualization}
\end{figure*}

\begin{figure*}[t]
  \centering
  \includegraphics[width=0.9\linewidth]{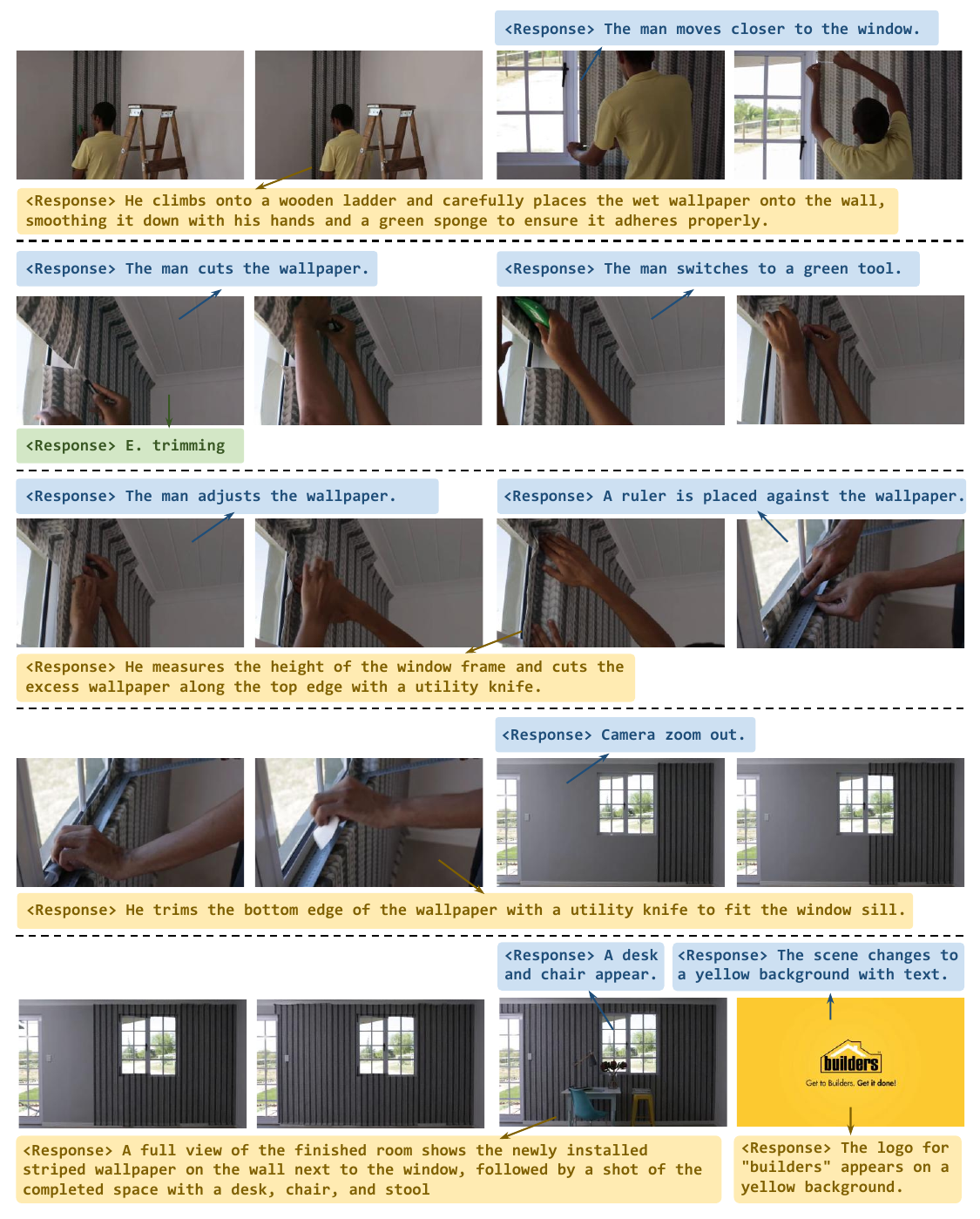}
  \caption{This is a continuation of the previous figure, showing the results for the same video.}
  \label{fig:visualization2}
\end{figure*}

In Fig.\ref{fig:visualization} and\ref{fig:visualization2}, we visualize the outputs of \modelname, which vividly illustrate its ability to interpret and appropriately respond even to instructions that were unseen during training. When confronted with task instructions that vary in both response granularity and content, the model consistently produces suitable outputs. These visualizations provide strong evidence that \modelname’s training framework successfully bridges the gap between offline model capabilities and the requirements of online streaming interactions, enabling reliable real-time responses that go far beyond simple QA.

\begin{table}[t]
\centering
\caption{Comparison of Existing Video Benchmarks. Streamo-Bench introduces the first mixed-task type specifically designed for streaming video.}
\label{tab:benchmark}
\resizebox{\linewidth}{!}{%
\begin{tabular}{@{}lcccc@{}}
\toprule
Benchmark     & \#Videos & \#Samples & Streaming & Task Type \\ \midrule
MVBench       & 3,673    & 4,000     & \color{red}{\XSolidBrush}         & QA   \\
TempCompass   & 410      & 7,540     & \color{red}{\XSolidBrush}         & QA   \\
ET-Bench      & 7,002    & 7,289     & \color{red}{\XSolidBrush}         & Mix  \\
SVBench       & 1,353    & 49,979    & \color{ForestGreen}{\CheckmarkBold}         & QA   \\
StreamBench   & 306      & 1,800     & \color{ForestGreen}{\CheckmarkBold}         & QA   \\
OVOBench      & 644      & 2,814     & \color{ForestGreen}{\CheckmarkBold}         & QA   \\
Streamo-Bench & 300      & 3000      & \color{ForestGreen}{\CheckmarkBold}         & Mix  \\ \bottomrule
\end{tabular}%
}
\end{table}

\subsection{Further Analysis of the Three-State Design}
To examine the rationale behind our training architecture more thoroughly, we compare the proposed Three-state Design with an alternative approach based on the [EOS] token. As shown in \cref{tab:eos}, the [EOS]-based model exhibits notable performance drops, particularly on proactive tasks (i.e., FAR) and grounding tasks. These results demonstrate that our three-state design consistently outperforms EOS-only training while introducing only negligible additional cost.

We attribute this gap to the fact that [EOS] maps both irrelevant and partially relevant segments to the same token. As a result, the model is encouraged to remain silent even when encountering relevant frames, causing it to miss the optimal timing for response. In contrast, the introduction of a [Standby] token alleviates this misalignment by explicitly marking relevant frames as soon as the event begins and preserving this state throughout the relevant interval. This leads to more accurate temporal alignment and more complete coverage, which is reflected in the higher grounding TIoU.

\begin{table}[h]
\centering
\setlength{\tabcolsep}{4pt}
\renewcommand{\arraystretch}{1.2}
\caption{Comparison on the same training dataset, Streamo-Instruct, where the only change is replacing the proposed three-state design with EOS-only training. Using only [EOS] degrades performance, especially on proactive prediction (FAR) and forward grounding, highlighting the benefit of the three-state design.}
\label{tab:eos}
\resizebox{\columnwidth}{!}{%
\begin{tabular}{l|cccc|c}
\hline
\multirow{2}{*}{Model} & \multicolumn{4}{c|}{OVOBench} & Streamo-Bench     \\ \cline{2-6} 
                       & RTVP  & BT    & FAR   & AVG   & Forward Grounding \\ \hline
Streamo-3B           & 61.51 & 41.76 & 53.72 & 52.33 & 14.7              \\
Streamo-3B w/ EOS    & 60.93 & 39.43 & 45.22 & 48.52 &  9.3           \\ \hline
\end{tabular}%
}
\end{table}

\begin{figure*}[t]
  \centering
  \includegraphics[width=0.9\linewidth]{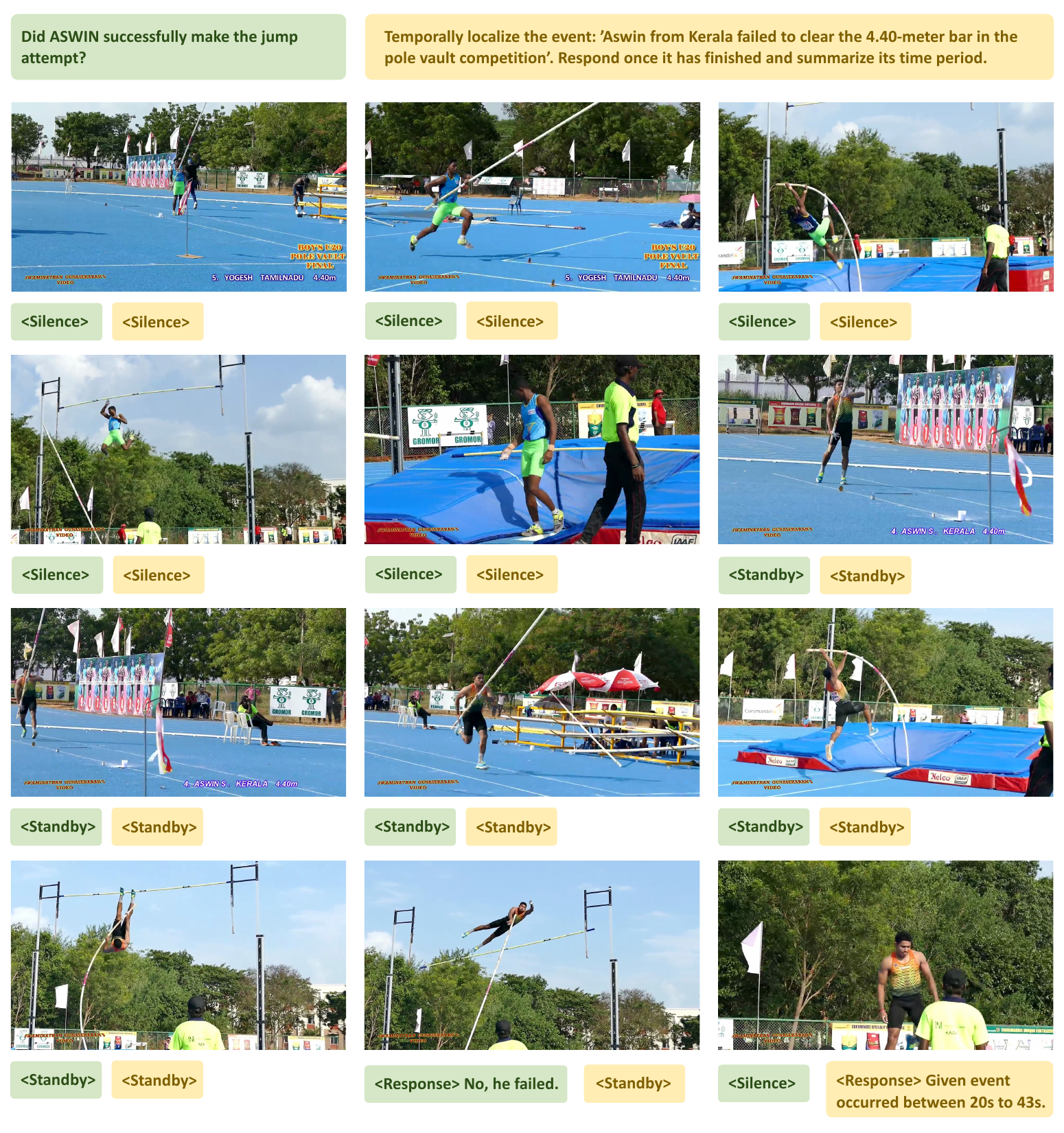}
  \caption{Visualization of the three-state decoding process. The model stays in \texttt{[Silence]} for irrelevant frames, switches to \texttt{[Standby]} once the query-relevant event involving ASWIN begins, and emits \texttt{[Response]} only after the outcome becomes clear. For the grounding task, the persistent \texttt{[Standby]} state helps preserve attention over the relevant interval and enables more complete temporal coverage of the event span.}
  \label{fig:standby}
\end{figure*}

A key advantage of [Standby] is that it explicitly models frames that are already relevant but not yet ready for a final response. As shown in \cref{fig:standby}, because the query is specifically about ASWIN, the model switches to [Standby] once ASWIN appears and the attempt becomes temporally relevant, even though the final outcome is still uncertain. This allows the model to preserve attention over the ongoing event instead of treating these frames as irrelevant. Meanwhile, for grounding, the continuous [Standby] state helps cover the full event span more completely, rather than activating only near the final decisive moment.

\begin{table*}[t]
  \centering
  \setlength{\tabcolsep}{4pt}
    \renewcommand{\arraystretch}{1.3}
      \caption{Performance of streamo compared to various MLLMs on ViSpeak-Bench.}
  \label{tab:ViSpeakBench}
  \resizebox{\linewidth}{!}{
      \begin{tabular}{lcccc|ccccccc|cccccccc|c}
        \hline
        \multirow{2}{*}{Method} & \multirow{2}{*}{Params} & \multirow{2}{*}{Frames} & \multirow{2}{*}{Omni} & \multirow{2}{*}{Streaming} & \multicolumn{7}{c|}{Time Accuracy (\%)} & \multicolumn{8}{c|}{Text Score}  &\multirow{2}{*}{Overall}  \\
        & & & &  & AW & VI & HR & VW & VT & GU  & \textbf{All}& VR & AW & VI & HR & VW & VT & GU  &\textbf{All} & \\
        \hline
        Human (Avg) & - & - & - & - & 70.00 & 100.00 & 90.00 & 92.00 & 96.00 & 98.80 & 91.13 & 4.80 & 2.45 & 4.58 & 3.06 & 5.00 & 5.00 & 2.85 & 3.96 & 3.69 \\
        Human (Max) & - & - & - & - & 70.00 & 100.00 & 100.00 & 100.00 & 100.00 & 100.00 & 95.00 & 5.00 & 2.71 & 5.00 & 3.62 & 5.00 & 5.00 & 3.19 & 4.22 & 4.01 \\
        \hline
        \multicolumn{21}{c}{\textbf{Proprietary MLLMs}}\\ \hline
        Gemini 1.5 pro~\cite{team2024gemini} & - & - & \checkmark & \ding{55} & 46.00 & 60.00 & 85.00 & 84.00 & 48.00 & 97.00 & 70.00 & 3.03 & 2.34 & 2.93 & 1.36 & 4.66 & 4.68 & 2.07 & 3.01 & 2.19 \\
        GPT-4o~\cite{hurst2024gpt4o} & - & - & \checkmark & \ding{55} & 48.50 & 82.00 & 96.00 & 99.00 & 100.00 & 99.50 & 87.50 & 3.18 & 2.27 & 3.53 & 1.71 & 5.00 & 4.98 & 2.22 & 3.27 & 2.99 \\
        \hline
        \multicolumn{21}{c}{\textbf{Open-Source Video MLLMs}}\\ \hline
        InternVL-2.5~\cite{chen2024internvl-2.5} & 8B & 16 & \ding{55} & \ding{55} & 41.50 & 55.50 & 46.00 & 96.00 & 72.00 & 99.50 & 68.42 & 2.93 & 2.16 & 3.67 & 0.74 & 3.05 & 4.81 & 1.26 & 2.66 & 1.98 \\
        Qwen2.5-VL~\cite{bai2025qwen25vl} & 7B & 1 fps & \ding{55} & \ding{55} & 42.50 & 78.00 & 31.00 & 95.00 & 85.00 & 98.50 & 71.67 & 2.34 & 2.31 & 2.31 & 1.32 & 5.00 & 3.91 & 1.02 & 2.60 & 2.25 \\
        Qwen2.5-VL~\cite{bai2025qwen25vl} & 72B & 1 fps & \ding{55} & \ding{55} & 44.50 & 81.00 & 77.00 & 91.00 & 91.00 & 93.00 & 79.58 & 3.15 & 2.64 & 3.36 & 1.00 & 5.00 & 5.00 & 1.50 & 3.09 & 2.62 \\
        VITA 1.5~\cite{vita} & 7B & 1 fps & \checkmark & \ding{55} &  18.00 & 46.00 & 40.00 & 88.00 & 49.00 & 97.50 & 56.42 & 2.40 & 2.08 & 0.57 & 0.85 & 4.57 & 4.49 & 1.18 & 2.31 & 1.54 \\
        Ola~\cite{ola} & 7B & 1 fps & \checkmark & \ding{55} & 27.00 & 67.00 & 44.00 & 89.00 & 69.00 & 98.50 & 65.75 & 2.95 & 1.81 & 2.67 & 0.55 & 4.71 & 3.67 & 1.52 & 2.55 & 1.86 \\
        FlashVstream~\cite{zhang2024flashvstream} & 7B & 1 fps & \ding{55} & \checkmark & 34.00 & 16.00 & 48.00 & 75.00 & 33.00 & 99.50 & 50.92 & 1.75 & 1.63 & 1.31 & 0.67 & 4.88 & 4.61 & 0.70 & 2.22 & 1.24 \\
        Dispider~\cite{qian2025dispider} & 7B & 16 & \ding{55} & \checkmark  & 38.50	&70.00	&44.00	&69.00&	100.00	&99.50	&70.17& 2.50&	1.75	&4.06	&0.91&	0.61	&2.49&	2.07	&2.06 &1.63\\
        ViSpeak~\cite{vispeak} & 7B & 1 fps & \checkmark & \checkmark & 56.50 & 72.00 & 83.00 & 93.00 & 79.00 & 99.00 & 80.42 & 3.75 & 2.63 & 3.84 & 1.07 & 4.95 & 3.15 & 3.36 & 3.25 & 2.76 \\
        \rowcolor{SkyBlue!20} Streamo & 7B & 1 fps & \ding{55} & \checkmark & 59.00 & 79.00 & 82.00 & 97.00 & 86.00 & 100 & 83.83 & 2.73 & 2.31 & 3.62 & 1.33 & 4.96 & 3.62 & 2.97 & 3.08 & 2.71 \\
        \hline
      \end{tabular}
    }
\end{table*}

\section{Streamo-Instruct}
\subsection{Data Generation Prompt}
We next elaborate on the prompts used in our data annotation pipeline. For event caption tasks, we leverage ARC-Hunyuan\cite{ge2025arc}, which is specifically trained for video segmentation and grounding, and directly adopt its official prompt for initial data processing. We then use the prompt in Tab.\ref{prompt:recap} to rewrite and clean the annotated caption sentences. For narration generation, which describes inter-frame temporal changes, the generation prompt is given in Tab.\ref{prompt:description}, and the prompt for merging and cleaning the resulting descriptions is provided in Tab.\ref{prompt:narration}. For the TSQA task, the detailed prompt is presented in Tab.~\ref{prompt:TSQA}.

\section{Streamo-Bench}
In Tab.~\ref{tab:benchmark}, we compare our proposed Streamo-Bench with existing video benchmarks. Streamo-Bench is, to the best of our knowledge, the first streaming video benchmark that integrates multiple task types. Existing streaming video benchmarks typically use QA as the sole evaluation task, which mainly measures perceptual understanding rather than the ability to perform diverse open-ended tasks. However, the ability to follow varied instructions and complete multiple tasks is a key requirement for streaming video models. By filling this gap, Streamo-Bench enables more comprehensive evaluation of a model’s instruction-following ability in open-ended streaming scenarios.

\subsection{Statistics}
Our benchmark contains 300 videos sampled from COIN\cite{tang2019coin}, YouCookv2\cite{zhou2018youcook2}, and ActivityNet~\cite{caba2015activitynet}. Each video is annotated with multiple tasks, including Grounding, Narration, Caption, and Time-Sensitive QA, yielding a total of 3{,}000 task-specific instances. Each video in Streamo-Bench contains 2x grounding (forward + backward) tasks, 1x dense caption task, and 1x narration task, with the rest being TSQA. This comprehensive design enables a thorough examination of a model’s ability to process and respond to diverse instructions in streaming settings.

\subsection{Metric}
To comprehensively evaluate the performance of models on our Streamo-Bench, we detail the metrics used for each task type below.  

\noindent
\textbf{Grounding Evaluation.}
For grounding tasks, we distinguish between forward (queries referring to time points before an event) and backward (queries referring to time points after an event) contexts. Performance is measured using mean Intersection over Union (mIoU), which quantifies the overlap between the model’s predicted temporal interval and the ground-truth interval.

Let the predicted and ground-truth temporal intervals, $t^{\mathrm{pred}}$ and $t^{\mathrm{gt}}$, for sample $i$ be:
\begin{equation}
t_i^{\mathrm{pred}} = [\,s_i^{\mathrm{pred}},\, e_i^{\mathrm{pred}}\,], \qquad
t_i^{\mathrm{gt}}   = [\,s_i^{\mathrm{gt}},\, e_i^{\mathrm{gt}}\,],
\end{equation}
where $s$ and $e$ represent the start and end timestamps, respectively. The IoU for sample $i$ is defined as the ratio of intersection length to union length:
\begin{equation}
\mathrm{IoU}_i =
\frac{\max\!\bigl(0,\, \min(e_i^{\mathrm{pred}}, e_i^{\mathrm{gt}}) - \max(s_i^{\mathrm{pred}}, s_i^{\mathrm{gt}})\bigr)}
{\max(e_i^{\mathrm{pred}}, e_i^{\mathrm{gt}}) - \min(s_i^{\mathrm{pred}}, s_i^{\mathrm{gt}})}.
\end{equation}

The mean IoU (mIoU) over $N$ samples is
\begin{equation}
\mathrm{mIoU} = \frac{1}{N}\sum_{i=1}^{N} \mathrm{IoU}_i.
\end{equation}

\noindent
\textbf{Narration and Caption Evaluation.}
Because narration and captioning are open-ended generation tasks, directly evaluating output quality is challenging. Following the evaluation protocol of Chatbot Arena \cite{zheng2023judging} and StreamingVLM \cite{xu2025streamingvlm}, we assess narration and caption quality via pairwise comparison against a strong baseline, Qwen2.5-VL-72B~\cite{bai2025qwen25vl}. The win rate is defined as the proportion of cases in which our model’s output is judged superior to the baseline’s output.

\noindent
\textbf{Time-Sensitive QA Evaluation.}
For Time-Sensitive QA, we require that a prediction be correct in both its content and its timestamp. Let $Q$ be the set of TSQA questions. For each question $q \in Q$, the ground truth consists of $m_q$ time-stamped answers:
\begin{equation}
G_q = \{(a_i^q, t_i^q)\}_{i=1}^{m_q},
\end{equation}
where $a_i^q$ is the answer content and $t_i^q$ is its timestamp. The model produces $n_q$ predictions:
\begin{equation}
P_q = \{(\hat{a}_j^q, \hat{t}_j^q)\}_{j=1}^{n_q},
\end{equation}
where $\hat{a}_j^q$ is the predicted content and $\hat{t}_j^q$ is the predicted timestamp.

A predicted pair $(\hat{a}_j^q, \hat{t}_j^q)$ may match a ground-truth pair $(a_i^q, t_i^q)$ only if it is correct in both content and time. For the content evaluation:
\begin{equation}
C(\hat{a}_j^q, a_i^q) =
\begin{cases}
1, & \text{if content matches},\\
0, & \text{otherwise}.
\end{cases}
\end{equation}
For the timestamp, we define a non-negative tolerance parameter $\delta_t \ge 0$. Then we evaluate the correctness of the timestamp by:
\begin{equation}
T(\hat{t}_j^q, t_i^q; \delta_t) =
\begin{cases}
1, & \text{if}\quad |\hat{t}_j^q - t_i^q| \le \delta_t,\\
0, & \text{otherwise}.
\end{cases}
\end{equation}

In our experimental setting, the $\delta_t$ is set to 3 seconds. For the $i$-th answer point of question $q$, we define an indicator $I_i^q$ that checks whether there exists at least one prediction satisfying both content and temporal constraints: :  
\begin{equation}
I_i^q = 
\begin{cases} 
1 & \text{if}\quad C(\hat{a}_j^q, a_i^q) = 1 \land T(\hat{t}_j^q, t_i^q; \delta_t) = 1 \\
0 & \text{otherwise}
\end{cases}
\end{equation}

The final accuracy and recall can be given as:  
\begin{equation}
\text{Accuracy} = \frac{1}{\sum_{q \in Q} m_q} \sum_{q \in Q} \sum_{i=1}^{m_q} I_i^q
\end{equation}

\begin{equation}
\text{Recall} = \frac{1}{|Q|} \sum_{q \in Q} \left( \frac{1}{m_q} \sum_{i=1}^{m_q} I_i^q \right)
\end{equation}

\subsection{Sample Visualization}
\begin{figure*}[t]
  \centering
  \includegraphics[width=0.9\linewidth]{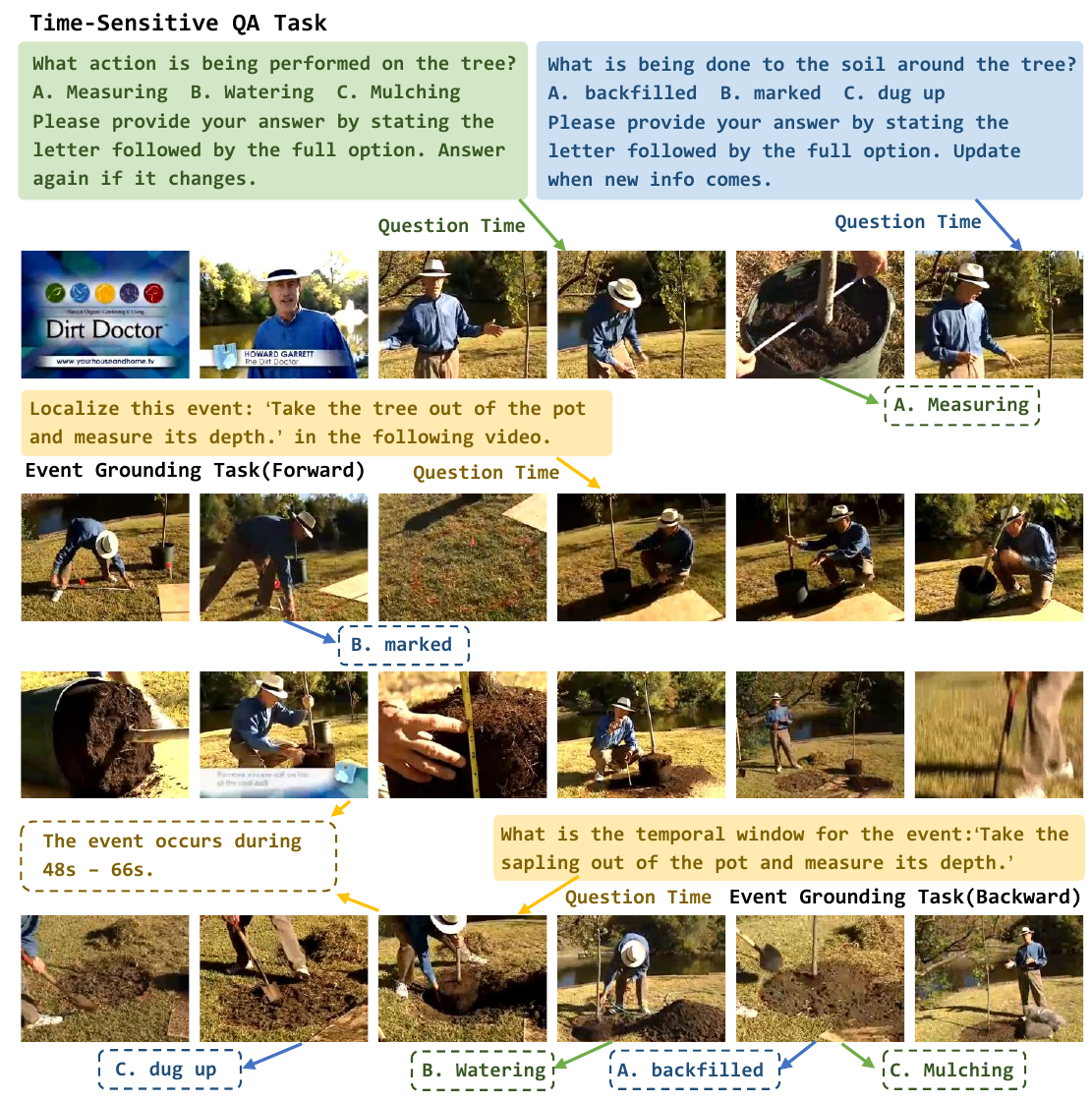}
  \caption{Streamo-Bench example illustrating multi-task instruction-following evaluation.}
  \label{fig:benchmark}
\end{figure*}

A sample instance from Streamo-Bench is illustrated in Fig.~\ref{fig:benchmark}. Forward and backward grounding questions are randomly placed either before or after their corresponding target temporal intervals. The TSQA question is inserted before the first answer timestamp. Narration and event caption instructions are placed before the start of the video stream to capture the overall video content.

\subsection{Further Analysis}
We further analyze the performance of existing models on Streamo-Bench and observe that their primary failures stem from a lack of instruction–task comprehension: they struggle to distinguish different task types and to produce task-appropriate outputs. This limitation arises because these models are typically trained exclusively on captioning or QA data, which constrains them to generate outputs tailored to only those specific tasks.

Examples in Tab.~\ref{tab:streamingvlm} clearly illustrate this phenomenon: while the models can satisfy caption or narration requirements, they often fail to understand grounding instructions and instead fall back to generic video descriptions. For TSQA tasks, although models trained on QA data can answer content-related questions, they do not properly follow instructions that require real-time updates to answers over the video timeline, leading to task failure.

In summary, existing models generally lack robust multi-task understanding, whereas Streamo-Bench is specifically designed to evaluate a model’s ability to interpret and respond to task-specific instructions in streaming scenarios.

\begin{table*}[t]
\centering
\renewcommand{\arraystretch}{1.2}
\setlength{\tabcolsep}{3pt}
\newcolumntype{C}[1]{>{\centering\arraybackslash}p{#1}}
\newcolumntype{L}[1]{>{\raggedright\arraybackslash}p{#1}}
\begin{tabular}{@{}C{0.18\linewidth}@{\hspace{6pt}}
                  C{0.18\linewidth}@{\hspace{6pt}}
                  C{0.18\linewidth}@{\hspace{6pt}}
                  C{0.18\linewidth}@{\hspace{6pt}}
                  C{0.18\linewidth}@{}}
\toprule
\multicolumn{5}{c}{\textbf{Input Video}} \\ \midrule
\includegraphics[width=\linewidth]{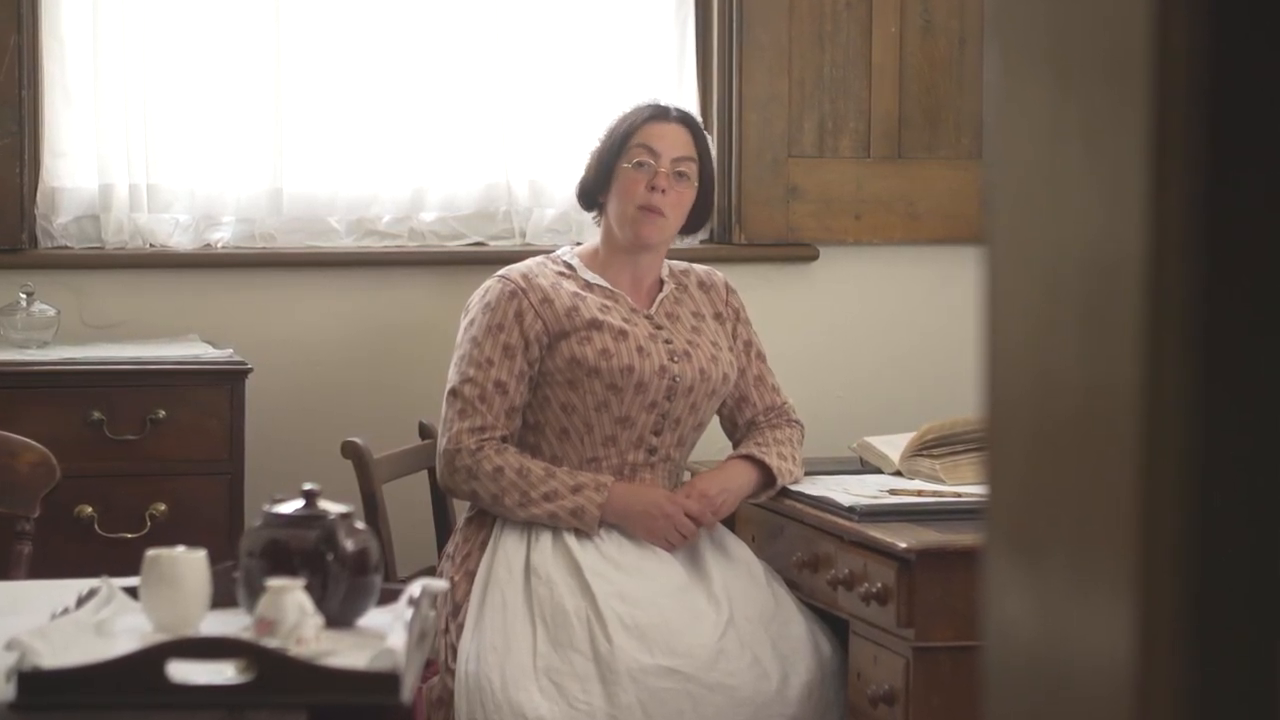} &
\includegraphics[width=\linewidth]{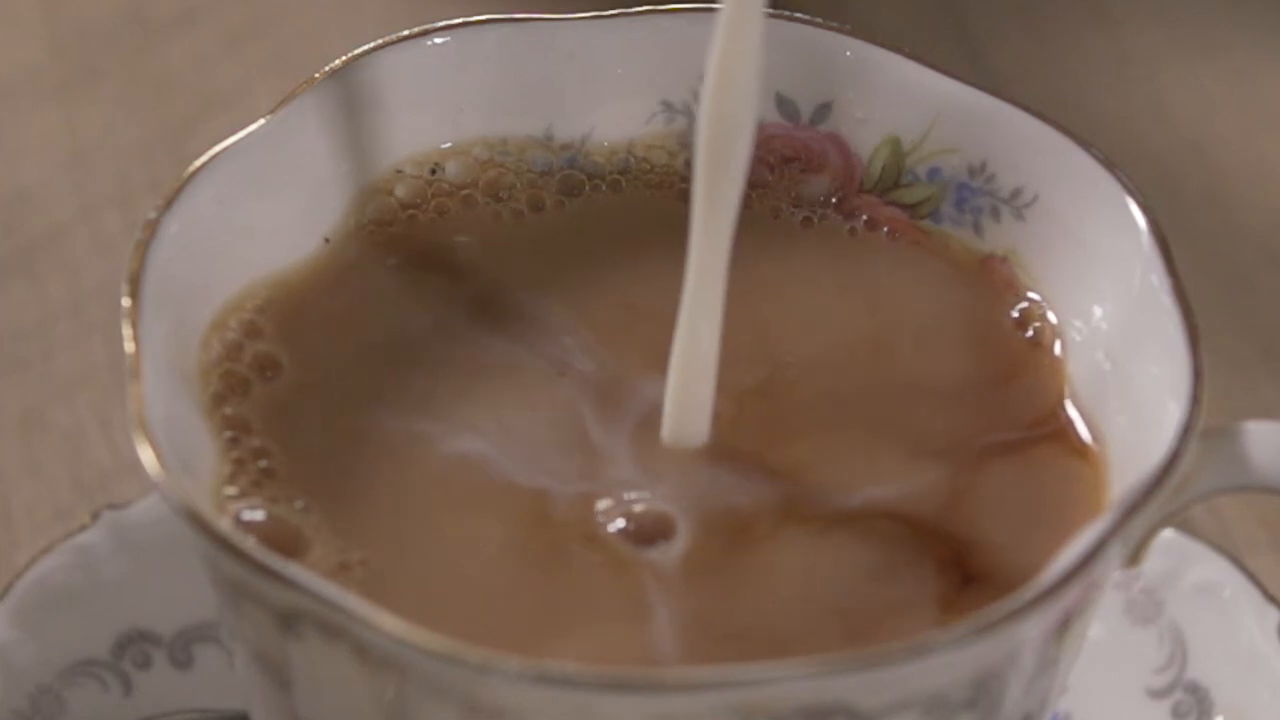} &
\includegraphics[width=\linewidth]{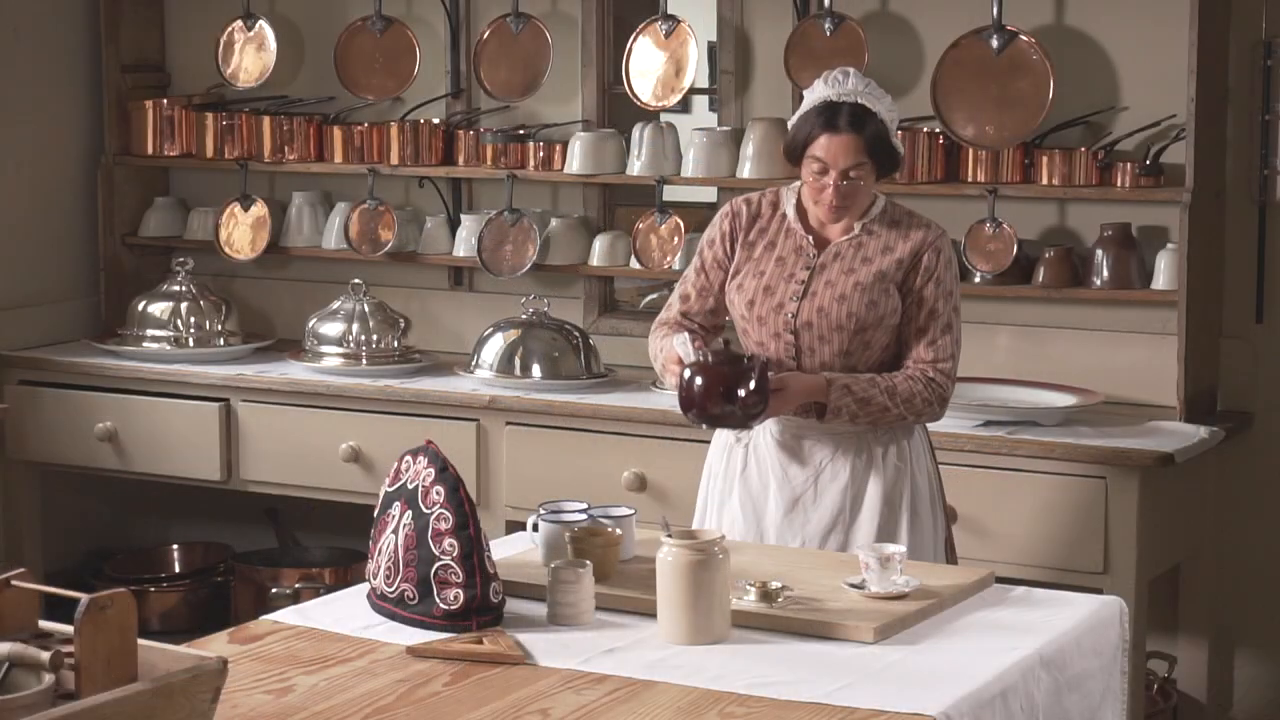} &
\includegraphics[width=\linewidth]{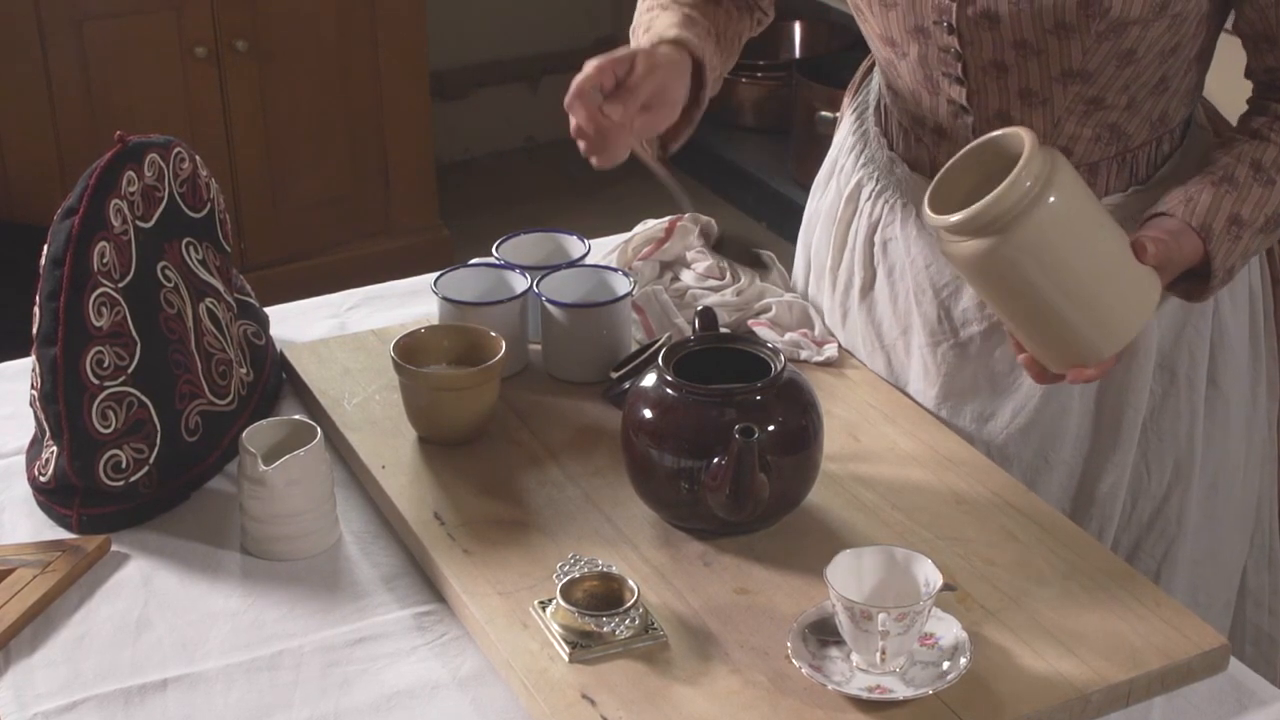} &
\includegraphics[width=\linewidth]{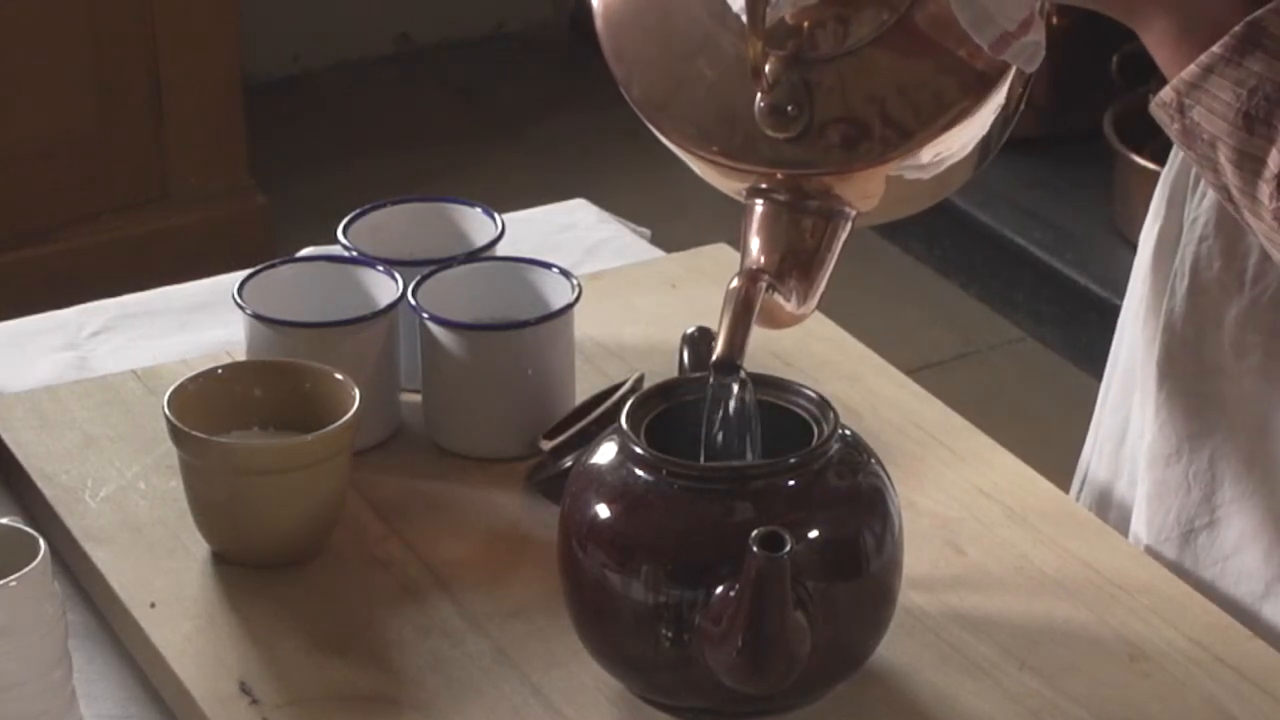} \\
\midrule
\multicolumn{2}{L{0.36\linewidth}}{\textbf{Prompt}} &
\multicolumn{3}{L{0.54\linewidth}}{\bfseries Model Output} \\
\cmidrule(lr){1-2}\cmidrule(lr){3-5}

\multicolumn{2}{L{0.36\linewidth}}{\textbf{Real-Time Narration Task:} Watch the following video and provide a real-time narration.} &
\multicolumn{3}{L{0.54\linewidth}}{Here's a typical mid-late 18th Century stoneware tea pot with the rounded spout and finial for pouring. After pouring when you were pouring out the last bit of liquid from the pot, you would be careful to stir away any bits of leaf which were floating around in the bottom.} \\ \midrule

\multicolumn{2}{L{0.36\linewidth}}{\textbf{Event Caption Task:} List every event happening in the video.} &
\multicolumn{3}{L{0.54\linewidth}}{We'll go through all the steps to make tea the way the Victorians did step by step. When you think of what they would be using here in their kitchen spaces, you can see they've got the teapot, and she's working on getting her leaves in. The method for both is basically exactly the same. With the fresh leaf version in this case, we're just adding in equal amounts of leaves and water.} \\ \midrule

\multicolumn{2}{L{0.36\linewidth}}{\textbf{Event Grounding Task:} Localize this event: ‘add some water to the tea’ in the following video.} &
\multicolumn{3}{L{0.54\linewidth}}{In the Victorian era the family kitchen was an even more important part of home life than in previous as this was where most of their time was spent. It was where the family ate their meals but also where they would come together. Whether it was afternoons at tea or a simple evening meal, the Victorian kitchen was a place that families would gather together and enjoy good company and food. Here we have made our own little tea service here, and I will be showing you how Mrs Bronte would serve it in her kitchen.} \\ \midrule

\multicolumn{2}{L{0.36\linewidth}}{\textbf{Time-Sensitive QA Task:} What is the woman doing? A. Preparing tea in kitchen B. Sitting at desk writing C. Drinking tea at desk. Please provide your answer by stating the letter followed by the full option. If the correct answer later changes, update your response.} &
\multicolumn{3}{L{0.54\linewidth}}{A. Preparing tea in kitchen. You will learn to make a cup of tea the Victorian way." To begin you'll need something sweet to top with your tea and milk. Tea is made with sugar or honey and the Victorians actually used a lot of cream for their tea. I'm going to use some milk for my tea today. And it can take up to two minutes to boil, so if you're making tea for more than one person.",  Start with just one cup at a time and then add more when ready.} \\ 

\bottomrule
\end{tabular}
\caption{Example output from StreamingVLM illustrating a failure to follow diverse task instructions and generate the corresponding response.}
\label{tab:streamingvlm}
\end{table*}

\begin{table*}
  \centering
  \resizebox{0.8\linewidth}{!}{
    \fcolorbox{black}{gray!20}{
    \parbox{0.8\linewidth}{
    \textbf{\textcolor{Green}{System Prompt}}:
        You are a helpful assistant specializing in streaming video analysis. \\
        You will receive input frame by frame, each labeled with absolute time intervals \\
        in the exact format \textless Xs-Ys\textgreater \  (e.g., \textless0s-1s\textgreater). Follow these rules precisely:\\
        \\
        1. Use \textless/Silence\textgreater \  when:\\
           \hspace*{1em}- No relevant event has started, OR \\
           \hspace*{1em}- The current input is irrelevant to the given question.  \\
        \\
        2. Use \textless/Standby\textgreater \  when:\\
           \hspace*{1em}- An event is in progress but has not yet completed, OR\\
           \hspace*{1em}- The current input is relevant but the question cannot yet be answered.  \\
        \\
        3. Use \textless/Response\textgreater \  only when:\\
           \hspace*{1em}- An event has fully concluded, OR\\
           \hspace*{1em}- The available information is sufficient to fully answer the question.\\
           Provide a complete description at this point.  \\
        \\
        Do not provide partial answers or speculate beyond the given information.  \\
        Whenever you deliver an answer, begin with \textless/Response\textgreater.
            }
        }
    }
    \caption{System prompt used in \modelname.}
    \label{prompt:streamo}
\end{table*}

\begin{table*}[t]
\centering
\setlength{\tabcolsep}{3pt}
\renewcommand{\arraystretch}{1.2}
\caption{Additional online benchmark evaluation results of Streamo framework with different base models (InternVL3 and Qwen3VL). Our framework consistently enables strong real-time streaming performance across diverse offline backbones.}
\label{tab:sup_online}
\resizebox{\textwidth}{!}{%
\begin{tabular}{@{}lccccccccccccccccc@{}}
\toprule
\multicolumn{1}{l|}{\multirow{2}{*}{Model}} & \multicolumn{1}{c|}{\multirow{2}{*}{\# Frames}} & \multicolumn{7}{c|}{Real-Time Visual Perception}                                                & \multicolumn{4}{c|}{Backward Tracing}                                   & \multicolumn{4}{c|}{Forward Active Responding}                          & Overall Avg. \\ \cmidrule(l){3-18} 
\multicolumn{1}{l|}{}                       & \multicolumn{1}{c|}{}                           & OCR   & ACR   & ATR   & STU   & FPD   & \multicolumn{1}{c|}{OJR}   & \multicolumn{1}{c|}{Avg.}  & EPM   & ASI   & \multicolumn{1}{c|}{HLD}   & \multicolumn{1}{c|}{Avg.}  & REC   & SSR   & \multicolumn{1}{c|}{CRR}   & \multicolumn{1}{c|}{Avg.}  & Overall Avg. \\ \midrule
\multicolumn{18}{c}{\textbf{Open-source Offline Models}} \\ \midrule
\multicolumn{1}{l|}{Qwen2-VL-72B~\cite{qwen2vl}}           & \multicolumn{1}{c|}{64}                         & 65.77 & \textbf{60.55} & \textbf{}69.83 & 51.69 & 69.31 & \multicolumn{1}{c|}{54.35} & \multicolumn{1}{c|}{61.92} & 52.53 & \textbf{60.81} & \multicolumn{1}{c|}{\textbf{57.53}} & \multicolumn{1}{c|}{\textbf{56.95}} & \textbf{38.83} & 64.07 & \multicolumn{1}{c|}{45}    & \multicolumn{1}{c|}{49.3}  & \textbf{56.27}        \\
\multicolumn{1}{l|}{LLaVA-Video-7B~\cite{llava-video}}         & \multicolumn{1}{c|}{64}                         & \textbf{69.13} & 58.72 & 68.83 & 49.44 & \textbf{74.26} & \multicolumn{1}{c|}{59.78} & \multicolumn{1}{c|}{63.52} & \textbf{56.23} & 57.43 & \multicolumn{1}{c|}{7.53}  & \multicolumn{1}{c|}{40.4}  & 34.1  & \textbf{69.95} & \multicolumn{1}{c|}{60.42} & \multicolumn{1}{c|}{\textbf{54.82}} & 52.91        \\
\multicolumn{1}{l|}{LLaVA-OneVision-7B~\cite{llava-onevision}}     & \multicolumn{1}{c|}{64}                         & 66.44 & 57.8  & \textbf{73.28} & \textbf{53.37} & 71.29 & \multicolumn{1}{c|}{\textbf{61.96}} & \multicolumn{1}{c|}{\textbf{64.02}} & 54.21 & 55.41 & \multicolumn{1}{c|}{21.51} & \multicolumn{1}{c|}{43.71} & 25.64 & 67.09 & \multicolumn{1}{c|}{58.75} & \multicolumn{1}{c|}{50.5}  & 52.74        \\
\multicolumn{1}{l|}{Qwen2-VL-7B~\cite{qwen2vl}}            & \multicolumn{1}{c|}{64}                         & 60.4  & 50.46 & 56.03 & 47.19 & 66.34 & \multicolumn{1}{c|}{55.43} & \multicolumn{1}{c|}{55.98} & 47.81 & 35.48 & \multicolumn{1}{c|}{56.08} & \multicolumn{1}{c|}{46.46} & 31.66 & 65.82 & \multicolumn{1}{c|}{48.75} & \multicolumn{1}{c|}{48.74} & 50.39        \\
\multicolumn{1}{l|}{InternVL-V2-8B~\cite{internvl}}         & \multicolumn{1}{c|}{64}                         & 67.11 & \textbf{60.55} & 63.79 & 46.07 & 68.32 & \multicolumn{1}{c|}{56.52} & \multicolumn{1}{c|}{60.39} & 48.15 & 57.43 & \multicolumn{1}{c|}{24.73} & \multicolumn{1}{c|}{43.44} & 26.5  & 59.14 & \multicolumn{1}{c|}{54.14} & \multicolumn{1}{c|}{46.6}  & 50.15        \\
\multicolumn{1}{l|}{LongVU-7B~\cite{longvu}}              & \multicolumn{1}{c|}{1fps}                       & 53.69 & 53.21 & 62.93 & 47.75 & 68.32 & \multicolumn{1}{c|}{59.78} & \multicolumn{1}{c|}{57.61} & 40.74 & 59.46 & \multicolumn{1}{c|}{4.84}  & \multicolumn{1}{c|}{35.01} & 12.18 & 69.48 & \multicolumn{1}{c|}{\textbf{60.83}} & \multicolumn{1}{c|}{47.5}  & 46.71        \\ \midrule
\multicolumn{18}{c}{\textbf{Open-source Online Models}}  \\ \midrule
\multicolumn{1}{l|}{Flash-VStream-7B~\cite{zhang2024flashvstream}}       & \multicolumn{1}{c|}{1fps}                       & 24.16 & 29.36 & 28.45 & 33.71 & 25.74 & \multicolumn{1}{c|}{28.8}  & \multicolumn{1}{c|}{28.37} & 39.06 & 37.16 & \multicolumn{1}{c|}{5.91}  & \multicolumn{1}{c|}{27.38} & 8.02  & 67.25 & \multicolumn{1}{c|}{60}    & \multicolumn{1}{c|}{45.09} & 33.61        \\
\multicolumn{1}{l|}{VideoLLM-online-8B~\cite{videollm-online}}     & \multicolumn{1}{c|}{2fps}                       & 8.05  & 23.85 & 12.07 & 14.04 & 45.54 & \multicolumn{1}{c|}{21.2}  & \multicolumn{1}{c|}{20.79} & 22.22 & 18.8  & \multicolumn{1}{c|}{12.18} & \multicolumn{1}{c|}{17.73} & -     & -     & \multicolumn{1}{c|}{-}     & \multicolumn{1}{c|}{-}     & -            \\
\multicolumn{1}{l|}{Dispider-7B~\cite{qian2025dispider}}               & \multicolumn{1}{c|}{1fps}                       & 57.72 & 49.54 & 62.07 & 44.94 & 61.39 & \multicolumn{1}{c|}{51.63} & \multicolumn{1}{c|}{54.55} & 48.48 & 55.41 & \multicolumn{1}{c|}{4.3}   & \multicolumn{1}{c|}{36.06} & 18.05 & 37.36 & \multicolumn{1}{c|}{48.75} & \multicolumn{1}{c|}{34.72} & 41.78        \\ \midrule
\multicolumn{18}{c}{\textbf{Streamo Framework}} \\ \midrule 
\multicolumn{1}{l|}{\MethodName-3B \textit{(Qwem2.5-VL)}}    & \multicolumn{1}{c|}{1fps}                       & 78.52 & 52.29 & 67.24 & 44.38 & 55.45 & \multicolumn{1}{c|}{71.20}  & \multicolumn{1}{c|}{61.51} & 51.18 & 57.43 & \multicolumn{1}{c|}{16.67} & \multicolumn{1}{c|}{41.76} & 27.94 & 50.72 & \multicolumn{1}{c|}{82.5}  & \multicolumn{1}{c|}{53.72} & 52.33        \\
\multicolumn{1}{l|}{\MethodName-7B \textit{(Qwem2.5-VL)}}    & \multicolumn{1}{c|}{1fps}                       & 79.19 & 57.80  & 75.00    & 49.44 & 64.36 & \multicolumn{1}{c|}{70.11} & \multicolumn{1}{c|}{65.98} & 54.55 & 52.03 & \multicolumn{1}{c|}{31.72} & \multicolumn{1}{c|}{46.10} & 29.96 & 51.03 & \multicolumn{1}{c|}{83.33} & \multicolumn{1}{c|}{54.77} & 55.61        \\
\rowcolor{SkyBlue!20}\multicolumn{1}{l|}{\MethodName-2B \textit{(InternVL3)}}   & \multicolumn{1}{c|}{1fps}      &  77.18     & 55.96    & 62.07     &  41.01     &  60.40     & \multicolumn{1}{c|}{70.11}      & \multicolumn{1}{c|}{61.12}      &  48.82     & 47.30      & \multicolumn{1}{c|}{13.44}      & \multicolumn{1}{c|}{36.52}      &  29.23     &  47.38     & \multicolumn{1}{c|}{80.42}      & \multicolumn{1}{c|}{52.34}      &  49.99            \\
\rowcolor{SkyBlue!20}\multicolumn{1}{l|}{\MethodName-4B \textit{(Qwen3-VL)}}      & \multicolumn{1}{c|}{1fps}     & 82.55      &  69.72     & 74.14      &  52.25     &  73.27     & \multicolumn{1}{c|}{81.52}      & \multicolumn{1}{c|}{72.24}      & 58.19      & 52.70      & \multicolumn{1}{c|}{17.20}      & \multicolumn{1}{c|}{42.70}      &  31.38     &  53.90     & \multicolumn{1}{c|}{84.17}      & \multicolumn{1}{c|}{56.48}      &  55.10           \\ \bottomrule
\end{tabular}%
}
\end{table*}

\begin{table*}[t]
\centering
\setlength{\tabcolsep}{4pt}
\renewcommand{\arraystretch}{1.2}
\caption{Additional offline benchmarks results of Streamo framework with different base models (InternVL3 and Qwen3VL). The results show that our training framework preserves the underlying offline capability while extending it to streaming video processing.}
\label{tab:sup_offline}
\resizebox{0.9\textwidth}{!}{%
\footnotesize
\begin{tabular}{@{}l|cccccc|c@{}}
\toprule
Model                           & \begin{tabular}[c]{@{}c@{}}OVO\\ Real-Time\end{tabular} & \begin{tabular}[c]{@{}c@{}}OVO\\ Backward\end{tabular} & MVBench & TempCompass & VideoMME & LongVideoBench & Avg \\ \midrule
\multicolumn{8}{c}{\textbf{Proprietary Models}}  \\ \midrule
Gemini-1.5-pro~\cite{team2024gemini} & 69.3 & 62.5 &	60.5	&	67.1	&	75.0	&	64.0 & 66.4 \\
GPT-4o~\cite{hurst2024gpt4o}	&     64.5 & 60.8   & 64.6	&	70.9	&	71.9	& 66.7	& 66.6 \\ \midrule
\multicolumn{8}{c}{\textbf{Open-source Online Models}}  \\ \midrule
Flash-VStream-7B~\cite{zhang2024flashvstream} & 28.4      & 27.4       & 61.2    & -        & 61.2     & -     & -      \\
VideoLLM-online-8B~\cite{videollm-online}    & 20.8      & 17.7       & 33.9    & -        & 26.9     & -     & -      \\
Dispider-7B~\cite{qian2025dispider}    & 54.6      & 36.1       & -    & -        & 57.2     & -     & -      \\
StreamingVLM-7B~\cite{xu2025streamingvlm}  & 62.0      & -       & 69.2    & -        & 65.1     & 59.0     & -      \\ \midrule
\multicolumn{8}{c}{\textbf{Streamo Framework}}  \\ \midrule
Qwen2.5-VL-3B ~\cite{bai2025qwen25vl}    & 54.6      & 37.8       & 67.0    & 64.4        & 61.5     & 54.2     & 56.6      \\
\rowcolor{SkyBlue!20}
\MethodName-3B       & 61.5 \color[HTML]{009901}(+6.9)  & 41.8 \color[HTML]{009901}(+4.0)  & 67.9 \color[HTML]{009901}(+0.9)   & 66.2 \color[HTML]{009901}(+1.8)       & 61.8 \color[HTML]{009901}(+0.3)    & 56.2 \color[HTML]{009901}(+2.0)      &  59.2 \color[HTML]{009901}(+2.6) \\ 
Qwen2.5-VL-7B ~\cite{bai2025qwen25vl}       & 58.8                & 42.2       & 69.6    & 71.7        & 65.1     & 56.0      &   60.6  \\
\rowcolor{SkyBlue!20}
\MethodName-7B   & 66.0 \color[HTML]{009901}(+7.2)               & 46.1 \color[HTML]{009901}(+3.9)      & 72.3 \color[HTML]{009901}(+2.7)   & 71.8     \color[HTML]{009901}(+0.1)   & 67.9 \color[HTML]{009901}(+2.8)    & 59.2 \color[HTML]{009901}(+3.2)    &  63.9 \color[HTML]{009901}(+3.3)   \\
InternVL3-2B ~\cite{zhu2025internvl3}                 & 59.5                & 36.4       & 70.4    & 57.6        & 58.9     & 55.4      & 56.4    \\
\rowcolor{SkyBlue!20}
Streamo-2B        & 61.1 \color[HTML]{009901}(+1.6)  & 36.5 \color[HTML]{009901}(+0.1)      & 71.4 \color[HTML]{009901}(+1.0)   & 57.8 \color[HTML]{009901}(+0.2)       & 60.1 \color[HTML]{009901}(+1.2)    & 56.5 \color[HTML]{009901}(+1.1)   &   57.3  \color[HTML]{009901}(+0.9)  \\ 
Qwen3-VL-4B~\cite{qwen3technicalreport}     & 66.5                & 42.8       & 68.9    & 65.8        & 69.3     & 53.2      &  61.1   \\
\rowcolor{SkyBlue!20}
Streamo-4B        & 72.2 \color[HTML]{009901}(+5.7)  & 42.7 \color[HTML]{CB0000}(-0.1)  & 70.4 \color[HTML]{009901}(+1.5)   & 66.3 \color[HTML]{009901}(+0.5)       & 68.7 \color[HTML]{CB0000}(-0.6)    & 56.1  \color[HTML]{009901}(+2.9)   &   62.8 \color[HTML]{009901}(+1.7)  \\ \bottomrule
\end{tabular}%
}
\end{table*}

\begin{table*}
  \centering
  \resizebox{0.8\linewidth}{!}{
    \fcolorbox{black}{gray!20}{
    \parbox{0.8\linewidth}{
    \textbf{\textcolor{Green}{Event Rewriting Prompt}}: You are given a set of video captions, each describing a specific moment in a video. For each caption, perform the following tasks:\\
\\
1. Remove any transition words, discourse markers, or sequence indicators (e.g., "Finally "Then "Next "Afterwards "At the beginning "At the end "The video ends with "The scene starts with etc.) at the beginning of the sentence or within the sentence, as these captions are now independent and do not need such connectors or structural descriptions.\\
2. Rewrite the caption to make it more concise and clear, without changing its meaning or omitting any important information.\\
3. Preserve all factual details and key actions described in the original caption.\\
4. Do not add any extra interpretation, information, or imagination not present in the original sentence. Only use the information given.\\
5. If the sentence includes a phrase describing the position of a shot or the sequence within the video (such as "The video ends with "At the start of the video "In the next scene "The video conclude with"), remove this part entirely. Focus only on describing the content of the shot.\\
\\
Example:\\
Original: "Finally, the video cuts back to the man in the indoor setting, who concludes the presentation by holding the bow."\\
Optimized: "The man in the indoor setting concludes the presentation by holding the bow."\\
\\
Process each caption in this way. Return the optimized sentence directly. \\
Original:\{sentences\} \\
Optimized:\\
        }
    }
  }
  \caption{Task prompt used for rewriting event caption.}
  \label{prompt:recap}
\end{table*}

\begin{table*}
  \centering
  \resizebox{0.8\linewidth}{!}{
    \fcolorbox{black}{gray!20}{
    \parbox{0.8\linewidth}{
    \textbf{\textcolor{Green}{Real-time Narration Task}}:\\
        \hspace*{1em}- Provide a continuous, time-synchronized narration of the video, describing actions, objects, and scene changes as they occur.\\
        \hspace*{1em}- Narrate the video in real time, updating the description frame-by-frame or moment-by-moment as events unfold.\\
        \hspace*{1em}- Generate live commentary of the video, focusing on who is doing what, where, and when, and noting any transitions or new events immediately.\\
        \hspace*{1em}- Deliver an on-the-fly description of the video, highlighting salient actions, interactions, and changes in context as soon as they appear.\\
        \hspace*{1em}- Produce a running narration that captures ongoing activities, brief pauses, and resumptions, maintaining temporal alignment with the video timeline.\\
    \textbf{\textcolor{Green}{Action Caption}}:\\
        \hspace*{1em}- Find, identify, and determine the temporal boundaries of a series of distinct actions or steps occurring throughout the video.\\
        \hspace*{1em}- Locate and describe a series of actions or steps in the video.\\
        \hspace*{1em}- Locate and pinpoint a sequential series of specific actions or steps in the video.\\
        \hspace*{1em}- Identify and mark the video segments corresponding to a series of actions or steps.\\
        \hspace*{1em}- Identify and localize a series of steps or actions occurring in the video.\\
    \textbf{\textcolor{Green}{Event Caption}}:\\
        \hspace*{1em}- Identify and describe all events in the following video.\\
        \hspace*{1em}- List every event happening in the following video with descriptions.\\
        \hspace*{1em}- Detect and summarize each event sequence in the following video.\\
        \hspace*{1em}- Extract and explain all notable events in the following video.\\
        \hspace*{1em}- Find all significant events in the following video and describe them.\\
    \textbf{\textcolor{Green}{Event Grounding}}:\\        
        \hspace*{1em}- Watch the following video and temporally localize the event. Respond once it has finished and summarize its time period. The given event is: '\{caption\}'\\
        \hspace*{1em}- Monitor the following video, identify the event, then respond after it finishes with a summary of its time window. The given event is: '\{caption\}'\\
        \hspace*{1em}- Analyze the following video, detect the event and report back upon its completion with its time period. The given event is: '\{caption\}'\\
        \hspace*{1em}- Review the following video, localize the event in time, then notify me once it ends and summarize the interval it occupies. The given event is: '\{caption\}'\\
        \hspace*{1em}- Identify and temporally segment the event in the following video. Report after it finishes with its time period and duration. The given event is: '\{caption\}'\\
    \textbf{\textcolor{Green}{ Time-sensitive QA}}:\\
        \hspace*{1em}- \{question\} If the answer changes over time, update your response accordingly.\\
        \hspace*{1em}- \{question\} Update your answer if it becomes different at a later time.\\
        \hspace*{1em}- \{question\} If it later differs, update your response promptly.\\
        \hspace*{1em}- \{question\} Refresh your answer upon any change.\\
        \hspace*{1em}- \{question\} If the correct answer later changes, update your response.
            }
        }
    }
    \caption{Prompt template used for diverse streaming video tasks.}
    \label{prompt:template}
\end{table*}

\begin{table*}[t]
  \centering
  \vspace*{-5mm}
  \resizebox{0.8\linewidth}{!}{
    \fcolorbox{black}{gray!20}{
    \parbox{0.8\linewidth}{
    \textbf{\textcolor{Green}{Video Description Prompt}}:
You are given two consecutive seconds in a video (2 frames per second). Please succinctly describe the most significant operation or change that occurred between these seconds, focusing on the following points:\\
1. Base your description solely on clearly observable information; avoid speculation or assumptions.\\
2. For each object or element that changed, briefly state what changed: position, movement, actions, shape, color, etc.\\
3. Only describe the main operation, event, or action that happened—avoid listing small movements or minor shifts.\\
4. Describe only the specific changed parts with clear and direct language; do not include unchanged content or summarize the overall scene.\\
5. Make your description short and focused, naming only the changes without referencing the sequence of frames or including explanations.\\
\\
Example:\\
'A woman appears.'\\
'You pick up a scissor.'\\
'The cup moves to the left.'\\
'A cat enters the frame.'\\
'The red ball rolls closer.'\\
'The lamp turns on.'\\
'The book closes.'\\
'A hand takes the remote.'\\
'The door opens further.'\\
\\
Only provide the most important description or a summary of multiple descriptions.
            }
        }
    }
    \caption{Task prompt used for frame-level video description generation.}
    \label{prompt:description}
    \vspace{-2mm}
\end{table*}

\begin{table*}[h]
  \centering
  \resizebox{0.8\linewidth}{!}{
    \fcolorbox{black}{gray!20}{
    \parbox{0.8\linewidth}{
    \textbf{\textcolor{Green}{Narration Generation Prompt}}:\\
**Objective**:\\
Clean the following second-by-second video descriptions to enhance coherence and eliminate redundancy. The original descriptions were generated with visibility of only the preceding and following 2 seconds, making them repetitive and disjointed.\\
\\
**Task**:\\
Transform the descriptions into a smooth, logical narrative by:\\
1. Removing Redundancy: Omit repeated descriptions of static or ongoing actions.\\
2. Filtering Insignificant Details: Exclude minor or fleeting actions that do not impact overall understanding.\\
3. Sentence Shortening: If a description significantly exceeds 5 words, rewrite it to approximately 5 words while preserving the main idea.\\
4. Merging Consecutive Events: Combine adjacent descriptions representing a continuous or complete action into a single, concise sentence (e.g., “002: Man touches socket” and “003: Socket disappears” → “003: Man removed socket”).\\
\\
**Output Format and Rules**:\\
1. Use the format: SSS: one-sentence description.\\
2. When merging or omitting descriptions, skip the corresponding timestamps.\\
3. Do not add explanations, notes, or blank lines.\\
4. If the descriptions are repetitive, monotonous, lack meaningful variation, or are confusing, ambiguous, or insufficient, output only: Negative Sample.\\
\\
Description:\\
\{Description\}
            }
        }
    }
    \caption{Task prompt used for merging the frame description to generate real-time narration.}
    \label{prompt:narration}
\end{table*}

\begin{table*}
  \centering
  \resizebox{0.8\linewidth}{!}{
    \fcolorbox{black}{gray!20}{
    \parbox{0.8\linewidth}{
    \textbf{\textcolor{Green}{TSQA Generation Prompt}}: You are a Time-Sensitive Video Question Generator. You need to identify all the elements in the video that change over time and formulate them into questions.\\
\\
**CORE REQUIREMENT**\\
Every question MUST have answers that CHANGE over time. If something doesn't change during the video, DO NOT create a question about it.\\
\\
**TASK**\\
\\
1. Identify ONLY aspects that visibly CHANGE during the video. Ignore:\\
\hspace*{1em} - Static elements that remain constant\\
\hspace*{1em} - Transitions, previews, close-ups that don't alter facts\\
\hspace*{1em} - Opening/closing sequences\\
\\
2. For each changing aspect, generate ONE question with MULTIPLE DIFFERENT answers:\\
\hspace*{1em} - Each question MUST have at least 2 DISTINCT answer values\\
\hspace*{1em} - Answers must represent actual changes observed at different times\\
\hspace*{1em} - Never repeat the same answer value\\
\\
3. Question types:\\
\hspace*{1em} - **Descriptive**: What/Which/Who (e.g., "What color is the ball?")\\
\hspace*{1em} - **Counting**: How many/How much (e.g., "How many people are visible?")\\
\hspace*{1em} - **State**: What stage (e.g., "What is the person doing?")\\
\hspace*{1em} - **Action** : What is being added/used (e.g., "What ingredient is being added?")\\
\hspace*{1em} - **Binary**: Yes/No (e.g., "Is the bacon cooked?")\\
\\
4. Answer format:\\
\hspace*{1em} - List answers chronologically\\
\hspace*{1em} - Include PRECISE time in seconds for each observed change\\
\hspace*{1em} - If state returns to a previous value, include it as a new entry\\
\\
**EXAMPLES**\\
{}[\{"question": "What color is the traffic light? "answers": [\{"value": "red "time": 3.8\}, \{"value": "green "time": 8.7\}, \{"value": "yellow "time": 23.2\}, \{"value": "red "time": 26.4\}]\},\\
\{"question": "How many people are in the frame? "answers": [\{"value": 1, "time": 0.0\}, \{"value": 2, "time": 3.8\}, \{"value": 3, "time": 17.1\}, \{"value": 1, "time": 42.6\}]\},\\
\{"question": "What is being poured into the glass? "answers": [\{"value": "water "time": 2.3\}, \{"value": "milk "time": 19.7\}, \{"value": "orange juice "time": 31.4\}]\},\\
\{"question": "Is the cake fully decorated? "answers": [\{"value": "No "time": 13.7\}, \{"value": "Yes "time": 48.9\}]\}{}]\\
\\
**OUTPUT**\\
\hspace*{1em} - Minimum 2 distinct answers per question\\
\hspace*{1em} - Precise timing for all changes. Times must reflect ACTUAL observed changes, not approximations\\
\hspace*{1em} - If unsure of exact timing, watch that segment again\\
\\
Think step-by-step and ensure all requirements are met and all time are precise.%
    }
    }
  }
  \caption{Task prompt used for Time-Sensitive QA generation.}
  \label{prompt:TSQA}
\end{table*}

\end{document}